\documentclass[pdflatex,sn-mathphys-num]{sn-jnl}% Math and Physical Sciences Numbered Reference Style
\usepackage{graphicx}%
\usepackage{multirow}%
\usepackage{amsmath,amssymb,amsfonts}%
\usepackage{amsthm}%
\usepackage{mathrsfs}%
\usepackage[title]{appendix}%
\usepackage{xcolor}%
\usepackage{textcomp}%
\usepackage{manyfoot}%
\usepackage{booktabs}%
\usepackage{algorithmicx}%
\usepackage{algpseudocode}%
\usepackage{listings}%
\usepackage{latexsym}
\usepackage[ruled,vlined,noline]{algorithm2e}
\usepackage{array}
\usepackage{hhline}
\usepackage{ulem}

\usepackage{url}
\usepackage{xcolor}
\usepackage{tabularx}
\usepackage{multirow}
\usepackage{makecell}
\usepackage{subcaption}
\usepackage{rotating}
\usepackage{colortbl}

\usepackage{hyperref}
\usepackage{cleveref}

\newcommand{\etal}{\textit{et al}.}

\definecolor{newcolor}{rgb}{.8,.349,.1}

\theoremstyle{thmstyleone}%
\theoremstyle{thmstyletwo}%

\theoremstyle{thmstylethree}%

\begin{document}

\title[Article Title]{Anatomy-Aware Text-Visual Fusion with Dual-Perspective Prompts for Fine-Grained Lumbar Spine Segmentation}
% \title[Article Title]{ATM-Net: Anatomy-Aware Text-Guided Multi-Modal Fusion for Fine-Grained Lumbar Spine Segmentation}


\author[1,3]{\fnm{Sheng} \sur{Lian}}
% \email{shenglian@fzu.edu.cn}

\author[1]{\fnm{Jianlong} \sur{Cai}}
% \email{jianlong.cai@foxmail.com}

\author[1]{\fnm{Dengfeng} \sur{Pan}}
% \email{p.df@foxmail.com}

\author[1,3]{\fnm{Guang-yong} \sur{Chen}}
% \email{gychen@fzu.edu.cn}

\author[1,3]{\fnm{Hao} \sur{Xu}}
% \email{241027081@fzu.edu.cn}

\author*[2]{\fnm{Fan} \sur{Zhang}}\email{zhangfan@sdtbu.edu.cn}

\author[2]{\fnm{Guodong} \sur{Fan}}
% \email{fanguodong@sdtbu.edu.cn}

\author[4]{\fnm{Jialun} \sur{Pei}}
% \email{jialunpei@cuhk.edu.hk}

\author[5]{\fnm{Shuo} \sur{Li}}
% \email{shuo.li11@case.edu}

\affil[1]{\orgdiv{College of Computer and Data Science}, \orgname{Fuzhou University}, \orgaddress{\city{Fuzhou}, \country{China}}}

\affil[2]{\orgdiv{School of Computer Science and Technology}, \orgname{Shandong Technology and Business University}, \orgaddress{\city{Yantai}, \country{China}}}

\affil[3]{\orgdiv{Engineering Research Center of Big Data Intelligence (Fuzhou University)}, \orgname{Ministry of Education (China)}, \orgaddress{\city{Fuzhou}, \postcode{350108}, \country{China}}}

\affil[4]{\orgdiv{Department of Computer Science and Engineering}, \orgname{The Chinese University of Hong Kong}, \orgaddress{\city{HKSAR}, \country{China}}}

\affil[5]{\orgdiv{Department of Biomedical Engineering and Computer and Data Science}, \orgname{Case Western Reserve University}, \orgaddress{\city{Cleveland}, \state{OH},\country{USA}}}

% \affil*[1]{\orgdiv{Department}, \orgname{Organization}, \orgaddress{\street{Street}, \city{City}, \postcode{100190}, \state{State}, \country{Country}}}

% \affil[3]{\orgdiv{Department}, \orgname{Organization}, \orgaddress{\street{Street}, \city{City}, \postcode{610101}, \state{State}, \country{Country}}}


\abstract{Accurate lumbar spine segmentation is crucial for diagnosing spinal disorders. Existing methods typically use coarse-grained segmentation strategies that lack the fine details needed for precise diagnosis. Additionally, their reliance on visual-only models hinders the capture of anatomical semantics, leading to misclassified categories and poor segmentation details. To address these limitations, we present ATM-Net, an innovative framework that employs an anatomy-aware, text-guided, multi-modal fusion mechanism for fine-grained segmentation of lumbar substructures, i.e., vertebrae (VBs), intervertebral discs (IDs), and spinal canal (SC). ATM-Net adopts the Anatomy-aware Text Prompt Generator (ATPG) to adaptively convert image annotations into anatomy-aware prompts in different views. These insights are further integrated with image features via the Holistic Anatomy-aware Semantic Fusion (HASF) module, building a comprehensive anatomical context. The Channel-wise Contrastive Anatomy-Aware Enhancement (CCAE) module further enhances class discrimination and refines segmentation through class-wise channel-level multi-modal contrastive learning. Extensive experiments on the MRSpineSeg and SPIDER datasets demonstrate that ATM-Net significantly outperforms state-of-the-art methods, with consistent improvements regarding class discrimination and segmentation details.
For example, ATM-Net achieves Dice of 79.39\% and HD95 of 9.91 pixels on SPIDER, significantly outperforming the competitive SpineParseNet by 8.31\% and 4.14 pixels, respectively.
The integration of anatomical text insights in ATM-Net represents a paradigm innovation in the community, providing clinicians with a powerful tool for precise diagnosis and treatment planning of lumbar spine disorders.}




\maketitle

% \Crefname{figure}[Fig.][Figs.]
% \Crefname{section}[Sec.][Secs.]
\Crefname{figure}{Fig.}{Figs.}   
\Crefname{figure}{Fig.}{Figs.}
\Crefname{section}{Sec.}{Secs.}   
\Crefname{section}{Sec.}{Secs.}
\Crefname{equation}{Eq.}{Eqs.}

%%%%%%%%%%%%%%%%%%% 1.介绍部分 %%%%%%%%%%%%%%%%%%%%%%%%%%%%%%%%%%%%%%
\section{Introduction}
\label{sec:intro}

%%%%%%%%%%%%%%%%%%% Fig摘要图 %%%%%%%%%%%%%%%%%%%%%%%%%%%%%%%%%%%%%%
\begin{figure*}[!t]
\centering
\includegraphics[width=1.05\textwidth,height=6.9cm]{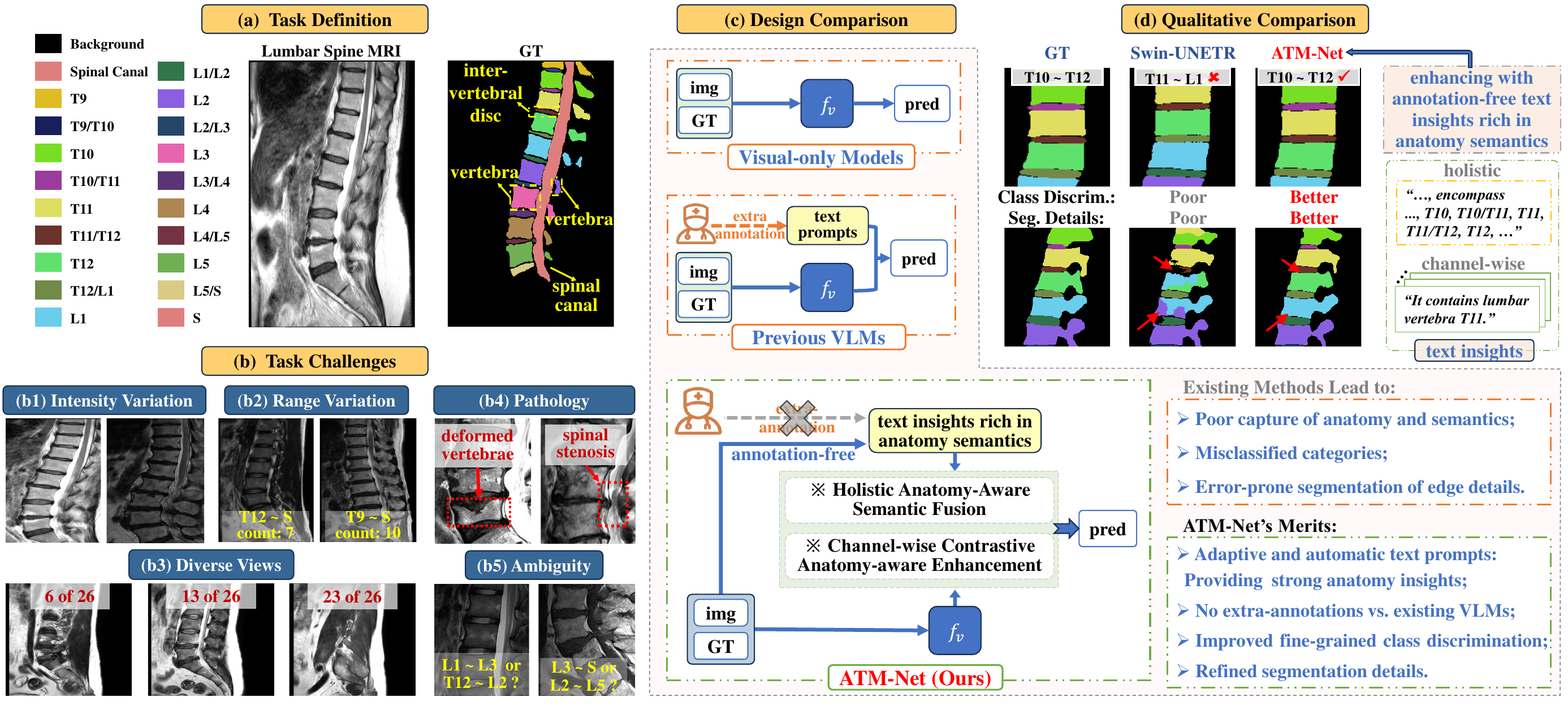}
\caption{(a) Task definition on the fine-grained lumbar spine MRI. (b) Task challenges in various aspects. (c) The design comparison between the visual-only models, the existing VLMs, and our ATM-Net. (d) ATM-Net's motivation in qualitative vies. Best viewed in color.} 
\label{fig:task-challenges}
\end{figure*}
%%%%%%%%%%%%%%%%%%% Fig摘要图 %%%%%%%%%%%%%%%%%%%%%%%%%%%%%%%%%%%%%%

% Please use \verb+elsarticle.cls+ for typesetting your paper.
% Additionally load the package \verb+medima.sty+ in the preamble using
% the following command: 
% \begin{verbatim} 
%   \usepackage{medima}
% \end{verbatim}

% Following commands are defined for this journal which are not in
% \verb+elsarticle.cls+. 
% \begin{verbatim}
%   \received{}
%   \finalform{}
%   \accepted{}
%   \availableonline{}
%   \communicated{}
% \end{verbatim}

% Any instructions relavant to the \verb+elsarticle.cls+ are applicable
% here as well. See the online instruction available on:
% \makeatletter
% \if@twocolumn
% \begin{verbatim}
%  http://support.stmdocs.com/wiki/
%  index.php?title=Elsarticle.cls
% \end{verbatim}
% \else
% \begin{verbatim}
%  http://support.stmdocs.com/wiki/index.php?title=Elsarticle.cls
% \end{verbatim}
% \fi
%%%%%%%%%%%%%%%%%%% 1.介绍部分 %%%%%%%%%%%%%%%%%%%%%%%%%%%%%%%%%%%%%%

Low back pain significantly impacts a wide range of patients' daily lives and work capabilities, posing significant challenges to healthcare systems~\citep{jenkins2025long}. 
Behind the low back pain issue lies complex lumbar disorders such as spondylolisthesis, lumbar disc herniation, and spinal stenosis, which are closely associated with lumbar spine substructures including vertebrae (VBs), intervertebral discs (IDs), and spinal canal (SC) ~\citep{berg2024machine,zhang2023lumbar,saremi2024evolution}. 
Accurate diagnosis and timely treatment of these issues are crucial for alleviating patient suffering and preventing disease progression, and MRI plays a pivotal role in this process~\citep{abel2023mri,foti2025spondylodiscitis}.
% MRI is essential for diagnosing lumbar spine conditions, offering non-invasive, high-resolution imaging that reveals detailed anatomical structures. 
Thus, the fine-grained multi-class segmentation of lumbar spine MRI (\Cref{fig:task-challenges}(a)), involving VBs, IDs, and SC, is a crucial step in diagnosing and treating lumbar spine-related disorders,
facilitating pathological localization~\citep{pang2022dgmsnet}, surgical planning~\citep{zhao2023attractive}, and therapeutic assessment~\citep{liang2024quantitative}.

However, the existing solutions typically adopt a coarse-grained segmentation strategy for the lumbar spine, falling short in nuanced diagnostics~\citep{bharadwaj2023deep,he2024lightweight}.
For example, ~\cite{he2023lsw,he2024lightweight,saenz2023automatic} developed segmentation models for lumbar MR images, categorizing all VBs, IDs, and the SC into three distinct classes. 
% Similarly, ~\cite{saenz2023automatic} proposed a variant of the U-Net architecture that can segment multiple anatomical structures, including vertebrae, intervertebral discs, and the spinal canal. However, this method also fails to achieve finer-grained segmentation (distinguishing individual vertebrae and intervertebral discs).
Compared to these solutions, achieving fine-grained segmentation in lumbar spine MRI presents challenges due to: (1) The images' diversity and complexity (\Cref{fig:task-challenges} (b1$\sim$b4)), and
(2) High similarity between the substructures (\Cref{fig:task-challenges} (b5)).
Hence, only a few solutions have been proposed for the fine-grained scenarios. 
% BianqueNet
Zheng \etal ~\citep{zheng2022deep} integrates three feature enhancement modules to segment 14 categories of lumbar substructures.
% ECSU-Net
% \cite{nazir2021ecsu} utilizes three-directional 2D subnetworks to enhance features collaboratively, thereby segmenting all vertebrae. 
Deng \etal ~\citep{deng2024effective} adopts a residual U-Net and BiSeNet hybrid network to achieve multi-class segmentation of vertebrae and intervertebral discs.

Despite promising progress, these visual-only models rely solely on visual features and struggle to capture the crucial anatomical semantics (Fig.\ref{fig:task-challenges}(c)). 
% They treat all pixels equally and cannot explicitly model the critical relationships between anatomical categories. 
They treat pixels without sufficient anatomical context, fail to explicitly model the critical relationships between substructures, and are often susceptible to factors such as class distribution imbalance. 
These limitations result in poor inter-class discrimination and errors in edge details (\Cref{fig:task-challenges}(d)).

% these visual-only models rely on the visual features of the images and struggle to accurately capture the crucial anatomical semantic information (\Cref{fig:task-challenges}(c,d)). 
% Specifically, these methods treat pixels without sufficient anatomical context and cannot explicitly model the critical relationships between substructures.

% This limitation results in poor inter-class discrimination and errors in edge details.
% Consequently, during fine-grained segmentation, they fall short in distinguishing between classes and are prone to errors in edge details.

Considering the advantages of large language models (LLMs), a challenge arises: \textit{\textbf{Can text information enhance fine-grained lumbar spine segmentation, and how can we efficiently extract and utilize these insights?}}
% Inspired by the advantages of large language models (LLMs), we leverage pre-trained biomedical language models to overcome the bottlenecks and bring new insights into this task.
This study aims to integrate text features rich in anatomical semantics, \textbf{offering notable benefits:}
It provides additional annotation-free text insights that inform the model, for example, that {\color{gray}\textit{T12} is above \textit{L1} and \textit{T12/L1} is between them}. Note that such text insights are robust to class distribution variations, effectively alleviate the decline in segmentation accuracy caused by class imbalance.
Specifically, unlike existing visual-language models (VLMs) that need additional expert annotations\cite{phan2024decomposing,li2024mlip}, we adaptively generate text prompts with rich anatomical semantics from image annotations.
This enables adaptive retrieval and deep exploration of anatomical semantics underlying segmentation annotations, enhancing inter-class discrimination and segmentation details.
However, integrating textual information into segmentation models, while advantageous, presents significant design challenges, including textual features extraction, fusion \& alignment of text-image information.

%%%%%%%%%%%%%%%%%%% 1.2方法总览部分 %%%%%%%%%%%%%%%%%%%%%%%%%%%%%%%%%%%%%%
% \subsection{Overall design of the proposed method}
%%%%%%%%%%%%%%%%%%% 1.2方法总览部分 %%%%%%%%%%%%%%%%%%%%%%%%%%%%%%%%%%%%%%

This study introduces \textbf{\textit{ATM-Net}}: an anatomy-aware, text-guided, multi-modal fusion framework for fine-grained lumbar spine segmentation.
% ATM-Net aims to conduct precisely fine-grained segmentation of vertebrae, intervertebral discs, and spinal canal from lumbar spine MRI, as illustrated in the left side of \Cref{fig:task-challenges}.
ATM-Net comprises three elaborately designed modules: The Anatomy-aware Text Prompt Generator (\textbf{ATPG}) module automatically and dynamically generates anatomy-aware text prompts in different views, incorporating information regarding anatomical structures and slice position, providing a solid foundation for the following modules.
The Holistic Anatomy-aware Semantic Fusion (\textbf{HASF}) module employs the multi-head attention and the cross-attention mechanisms to seamlessly integrate text and image features, building a comprehensive global semantic context for further model learning.
The Channel-wise Contrastive Anatomy-Aware Enhancement (\textbf{CCAE}) module, on the other hand, refines classification at the channel level by contrastive learning, enhancing the segmentation accuracy of specific anatomical structures.
Notably, the text features rich in semantic and anatomical information are extracted by large language models (LLMs) from text prompts generated by ATPG.
% In summary, ATM-Net facilitates effective collaboration among these modules by seamlessly integrating text features that are automatically generated with deep anatomical knowledge. This integration enhances the holistic representation of image features and boosts the identification of anatomical structures at the channel level. Consequently, it enriches both inter-class and intra-class semantic discrimination, thereby elevating ATM-Net's performance on the challenging task of fine-grained lumbar spine segmentation.
% We summarize the main contributions of ATM-Net as follows.
In summary, ATM-Net seamlessly integrates these modules to enhance image representation and anatomical identification, thereby significantly improving fine-grained lumbar spine segmentation.   

The main contributions of ATM-Net are as follows:
\begin{itemize}
    \item We propose ATM-Net, a innovative framework that integrates anatomy-aware text guidance for the fine-grained segmentation of various lumbar spine structures, e.g., VBs, IDs, and SC, offering a crucial tool for nuanced clinical diagnosis and treatment.
    % We propose ATM-Net, an innovative framework that utilizes anatomy-aware text insights to fine-grained segment the lumbar spine, enhancing nuanced diagnosis.
    \item ATPG adaptively converts image annotation into anatomy-aware text prompts in an annotation-free manner, with inherent robustness to class distribution variations that mitigate class imbalance in the task. These insights are further integrated with image features via HASF, enhancing ATM-Net's understanding of holistic anatomical context. Additionally, the CCAE module refines segmentation by improving inter-class discrimination and segmentation details through class-wise, channel-level contrastive learning.
    \item Across the MRSpineSeg and SPIDER datasets, extensive experiments demonstrate that ATM-Net consistently outperforms state-of-the-art solutions, including general medical image segmentation models, spine-specific approaches, and representative VLMs, in both class discrimination and segmentation details.
\end{itemize}

\section{Related work}

\subsection{Spine segmentation in MRIs}
% Research on spine segmentation in MRI has been limited, with three main factors preventing the development of spine segmentation algorithms. First, there is the challenge posed by inherent complexities in MRI spine segmentation. The water–fat interfaces, adjacent soft tissues like tendons, or focal air in the vicinity of the bone share the same intensity as (cortical) bone, leading to potential inaccuracies in segmentation results \citep{florkow2022magnetic}. Secondly, while CT segmentation algorithms benefit from a plethora of publicly accessible datasets with manually annotated masks, there are very few known datasets offering this essential resource in the MRI field \citep{ambellan2019automated,van2024lumbar}. Lastly, concerning economic viability, the per-examination cost is about 50\% higher for MRI than for CT \citep{mayerhoefer2020pet}. Moreover, MRI scans necessitate extended durations, which frequently result in prolonged waiting times for patients.

Spine segmentation in MRI is challenging due to similar appearances, image noise, and limited annotated datasets\citep{matos2023lumbar,  van2024lumbar, tian2025attention}.

% Research on spine segmentation in MRI is limited due to its inherent complexities, such as water-fat interfaces and adjacent soft tissues sharing intensities with bone, which can lead to inaccurate predictions~\citep{florkow2022magnetic}. Additionally, the publicly available datasets with manually annotated masks are insufficient~\citep{ambellan2019automated,van2024lumbar}. Economic factors also play a role, as MRI exams are costly and require longer scan times than CT, resulting in increased patient waiting times~\citep{mayerhoefer2020pet}.

In recent years, researchers have proposed different methods to improve the performance of MRI spinal segmentation. For instance,~\citep{deng2024effective,deng2023modified} developed an enhanced BiSeNet that integrates spatial features and multi-scale attention mechanisms, effectively enhancing the multi-class segmentation performance. Meanwhile, some studies have introduced novel techniques to address issues related to data annotation and class imbalance. The two-stage semi-supervised learning framework proposed by Huang \etal~\cite{huang2023semi} not only reduces the workload of data annotation but also optimizes the sample distribution in datasets. Pang \etal~\cite{pang2022dgmsnet} developed a hybrid network combining keypoint detection and segmentation prediction to further boost accuracy.
% To simplify the segmentation task, most studies tend to reduce the number of target categories, which can be detrimental to clinical diagnosis and treatment. 
Zhang \etal~\citep{zhang2024spinemamba} pioneered the application of the Mamba architecture to lumbar spine segmentation, demonstrating excellent performance in capturing long-range spatial relationships. However, their focus was primarily on vertebral segmentation. In contrast, Kim \etal~\citep{kim2024deep} leveraged CycleGAN to enhance the accuracy of intervertebral disc segmentation across MRIs from different manufacturers and with varying scanning parameters, yet their study was confined to disc segmentation. 

Our method introduces a lumbar spine segmentation method, achieving fine-grained segmentation of various VBs, IDs, and SC. We also integrate anatomical text insights for the first time, maximizing the use of existing annotated resources and generating additional anatomical knowledge to guide a more precise segmentation process.

\subsection{Textual insights for medical imaging tasks}
In the field of medical image segmentation, Visual Language Models (VLMs) offer promising solutions by combining techniques from the computer vision communities and the natural language processing communities \citep{lu2024visual}. 
This section briefs their two core components, i.e., \textit{text-guided medical image segmentation with VLMs}, and \textit{textual prompt engineering}.

\textit{\textbf{Text-guided medical image segmentation with VLMs.}}
Inspired by the success of large models in language processing 
\citep{devlin2019bert, yang2024harnessing, guo2025deepseek},
% schick2024toolformer, singhal2023large, luddecke2022image}, 
% \citep{devlin2019bert, brown2020language, singhal2023large, luddecke2022image}, 
VLMs have been applied to medical image segmentation \citep{li2023lvit, chen2024bi,guo2024common,rahman2025text}.
% \citep{li2023lvit, wang2022cris, park2022per}. 
Challenges in medical images, such as indistinct boundaries and minimal grayscale variations, make the direct application of natural image models impractical. 
Text-guided segmentation aims for pixel-level alignment between visual features and the textual prompts. Methods either use text insights for object recognition or fit cross-modal features through attention mechanisms 
\citep{zhang2024madapter, hu2024lga, huang2024adapting}. 
% \citep{feng2021encoder, kim2022restr, huang2020referring, yang2021bottom, yu2018mattnet, chen2019see, ding2021vision, feng2021encoder, hu2020bi, jing2021locate, kim2022restr, shi2018key, ye2019cross, zhou2023text}. 
Works like STPNet, TAV-M, VCLIPSeg, and CausalCLIPSeg have advanced alignment-based attention \citep{shan2025stpnet,rahman2025text,li2024vclipseg,chen2024causalclipseg} to integrate text insights.
% \citep{yang2022lavt, liu2023gres, liu2023polyformer}

\textit{\textbf{Textual prompt engineering.}}
In contrast to the elaborate annotations employed in visual tasks, the prior knowledge conveyed by class names is of a coarse granularity, posing challenges for precise multimodal alignment and downstream tasks.
Chen \etal~\citep{chen2023enhancing} suggested that meticulously designed medical prompts significantly enhance visual tasks.
Thus, textual prompt engineering has evolved, impacting areas like image classification, object detection, and image generation \citep{li2024promptad,zheng2024large,yun2025learning}.
% \citep{radford2021learning, shu2022test, hao2024optimizing, wang2022ofa}. 
In medical VLMs, MedPrompt automatically generates medical prompts for low-cost image classification \citep{zheng2024exploring}. Li \etal~\citep{li2024tp} focuses on enhancing diabetic retinopathy lesion segmentation by leveraging explicit textual prompts to guide the Segment Anything Model.
Liu \etal~\cite{liu2025synpo} focuses on boosting training-free few-shot medical image segmentation by improving the quality of negative prompts.
Tiu \etal~\citep{tiu2022expert} employs positive and negative textual prompts to enable zero-shot classification of pathologies in chest X-ray images.

% In medical VLMs, \citep{chen2023enhancing} identified effective prompt engineering techniques for medical applications. GloRIA generates clinically specific prompts, CheXzero uses binary prompts for disease classification, and MedKLIP enriches visual data with clinical descriptions \citep{huang2021gloria, tiu2022expert, wu2023medklip}.

We propose an automated pipeline to adaptively develop medical prompts in an annotation-free manner, highlighting semantic relationships between anatomical structures. By leveraging multi-level attention mechanisms and class-wise contrastive learning, we effectively integrate textual and visual features, ensuring efficient segmentation of substructures in lumbar spinal images.

\begin{figure*}[ht]
\centering
\includegraphics[width=1.05\textwidth,height=6.9cm]{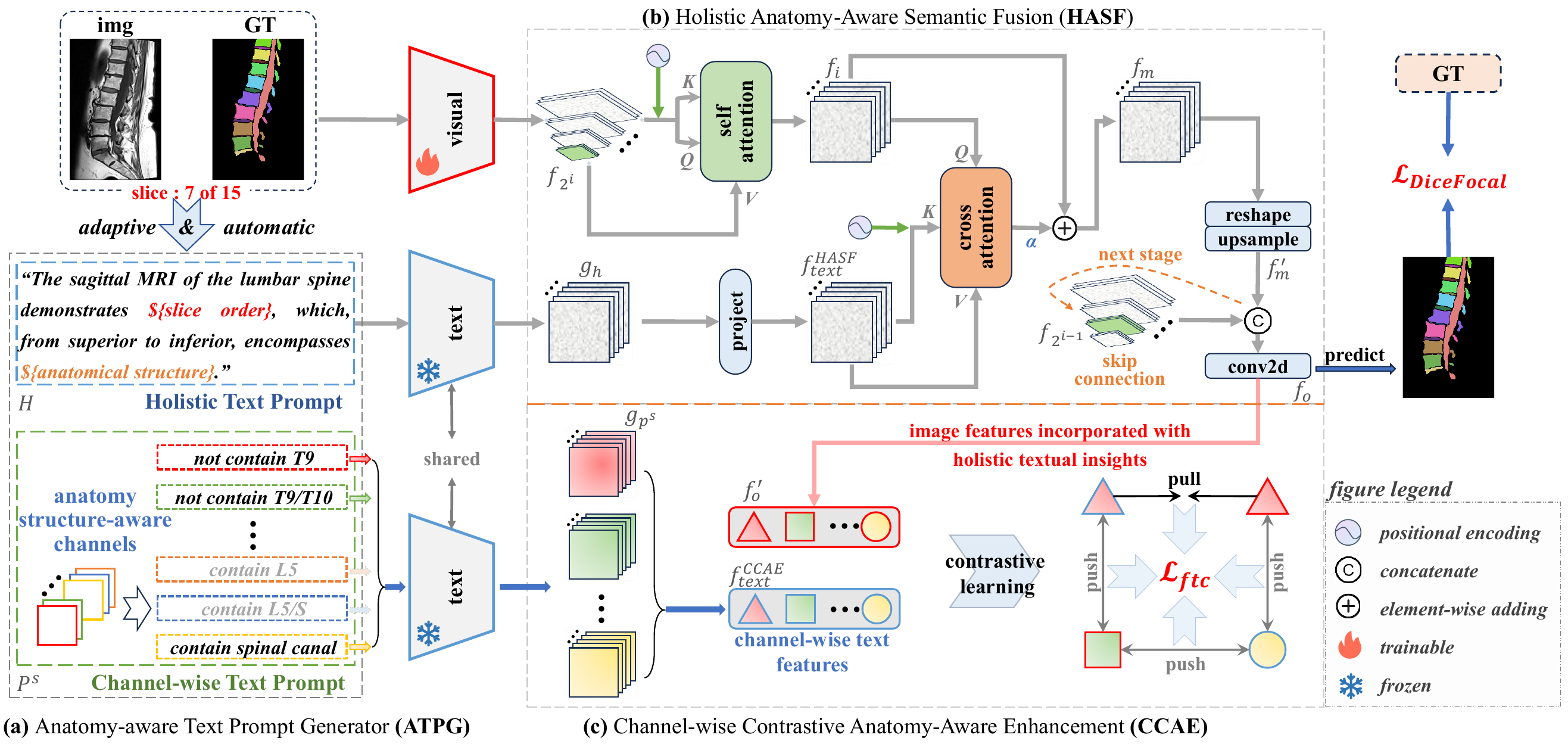}
\caption{\textbf{\textit{Method overview.}} 
ATM-Net integrates anatomy-aware text guidance for the fine-grained segmentation of lumbar spine substructures, which has three key components:
ATPG adaptively converts image annotation into anatomy-aware text prompts.
These insights are integrated with visual features via HASF, building a comprehensive anatomical context.
CCAE further enhances class discrimination \& segmentation details through class-wise channel-level multi-modal contrastive learning.
% HASF seamlessly fuses holistic text insights with image features across various scales.
% While CCAE utilizes contrastive learning to further enhance cross-modal knowledge complementarity regarding class discrimination and segmentation detail.
Best viewed in color.} 
\label{fig:framework}
\end{figure*}

%%%%%%%%%%%%%%%%%%% Fig主图 %%%%%%%%%%%%%%%%%%%%%%%%%%%%%%%%%%%%%%

\section{Methodology}

\subsection{Overview}

\textit{\textbf{Task definition.}}
The task of fine-grained lumbar spine image segmentation employs dataset $D = \{ (x_i, y_i) \}_{i=1}^{N}$ with $N$ annotated images.
Each image $x_i \in \mathbb{R}^{H \times W}$ has a corresponding segmentation mask $y_i \in \{0, 1, \ldots, C-1\}^{H \times W}$, covering fine-grained lumbar spine categories including various \textit{VBs, IDs, SC,} and the background (depicted in \Cref{fig:task-challenges}(a)).
Our primary objective is to leverage this dataset to train a model that can efficiently extract and integrate anatomy-aware textual information from image annotations, thereby delivering accurate and reliable segmentation outputs of various VBs, IDs, and SC, thus facilitating precise diagnosis and treatment for lumbar spine-related disorders.
% thus promoting improved diagnostic and therapeutic decisions.

% Our primary objective is to leverage this dataset to train a model that can deliver accurate and reliable segmentation outputs of various vertebrae, intervertebral discs, and spinal canal, thus facilitating improved diagnostic and therapeutic decisions.

\textbf{\textit{Method overview.}}
This paper presents an \textbf{A}natomy-aware \textbf{T}ext-guided \textbf{M}ulti-modal fusion image segmentation framework, termed \textbf{ATM-Net}, designed for achieving fine-grained multi-class segmentation of lumbar spine MRI.
ATM-Net introduces a pioneering automatic and adaptive approach that directly extracts anatomy-aware text prompts from annotated images and seamlessly integrates these critical insights with image information using LLMs.
Such mechanism is achieved by the following key components, including:
(1) The vision encoder and text encoder (\Cref{subsec:encoder}).
(2) ATPG: The anatomy-aware text prompt generator (\Cref{subsec:ATPG}). 
% which adaptively generates text prompts from labeled images  with anatomy awareness,
(3) HASF: The Holistic Anatomy-Aware Semantic Fusion module (\Cref{subsec:HASF}).
(4) CCAE: The Channel-wise Contrastive Anatomy-Aware Enhancement module (\Cref{subsec:CCAE}).
The overall design of ATM-Net is illustrated in \Cref{fig:framework}.

\subsection{The vision encoder and text encoder}
\label{subsec:encoder}

ATM-Net utilizes \textit{Swin-UNETR}~\citep{tang2022self} as its vision encoder, an advanced feature extractor well-suited for medical image analysis tasks.
For an input image $x_i \in \mathbb{R}^{H \times W \times 1}$, we extract multi-scale feature maps from various stages of Swin-UNETR, including 
$f_{2^i} \in \mathbb{R}^{\frac{H}{2^i} \times \frac{W}{2^i} \times C_i}, (i \in [1,\dots, 5])$.
% $f_2 \in \mathbb{R}^{\frac{H}{2} \times \frac{W}{2} \times C_1}$,
% $f_4 \in \mathbb{R}^{\frac{H}{4} \times \frac{W}{4} \times C_2}$,
% $f_8 \in \mathbb{R}^{\frac{H}{8} \times \frac{W}{8} \times C_3}$,
% $f_{16} \in \mathbb{R}^{\frac{H}{16} \times \frac{W}{16} \times C_4}$, and
% $f_{32} \in \mathbb{R}^{\frac{H}{32} \times \frac{W}{32} \times C_5}$,
Here, $C_i$ represents the channel dimensions at stage $i$, and $H$ and $W$ correspond to the original height and width of the input image.

To encode textual information, we adopt \textit{Bio\_ClinicalBERT}, a pre-trained LLM specifically designed for the biomedical domain~\citep{alsentzer2019publicly}.
Given textual prompts enriched with anatomical semantic information, including holistic prompts  $H \in \mathbb{R}^L$, and channel-specific textual prompts ${P^s} \in \mathbb{R}^T$,$(s \in [0,\dots, 19])$ (see \Cref{subsec:ATPG}),
Bio\_ClinicalBERT generates the corresponding text features $g_h \in \mathbb{R}^{L \times C}$
and $g_{p^s} \in \mathbb{R}^{T \times C}$, respectively.
Here, $C$ is the channel dimensions, while $L$ and $T$ denote the lengths of the holistic and channel-specific textual prompts, respectively. Note that $s$ represents the specific category encoding, and the total number of categories in this study is set to 20 (see \Cref{tab:dataset_characteristics}).

\subsection{Anatomy-aware Text Prompt Generator (ATPG)}
\label{subsec:ATPG}

\begin{figure}[h]
    \centering
    \includegraphics[width=0.92\linewidth,height=6.3cm]{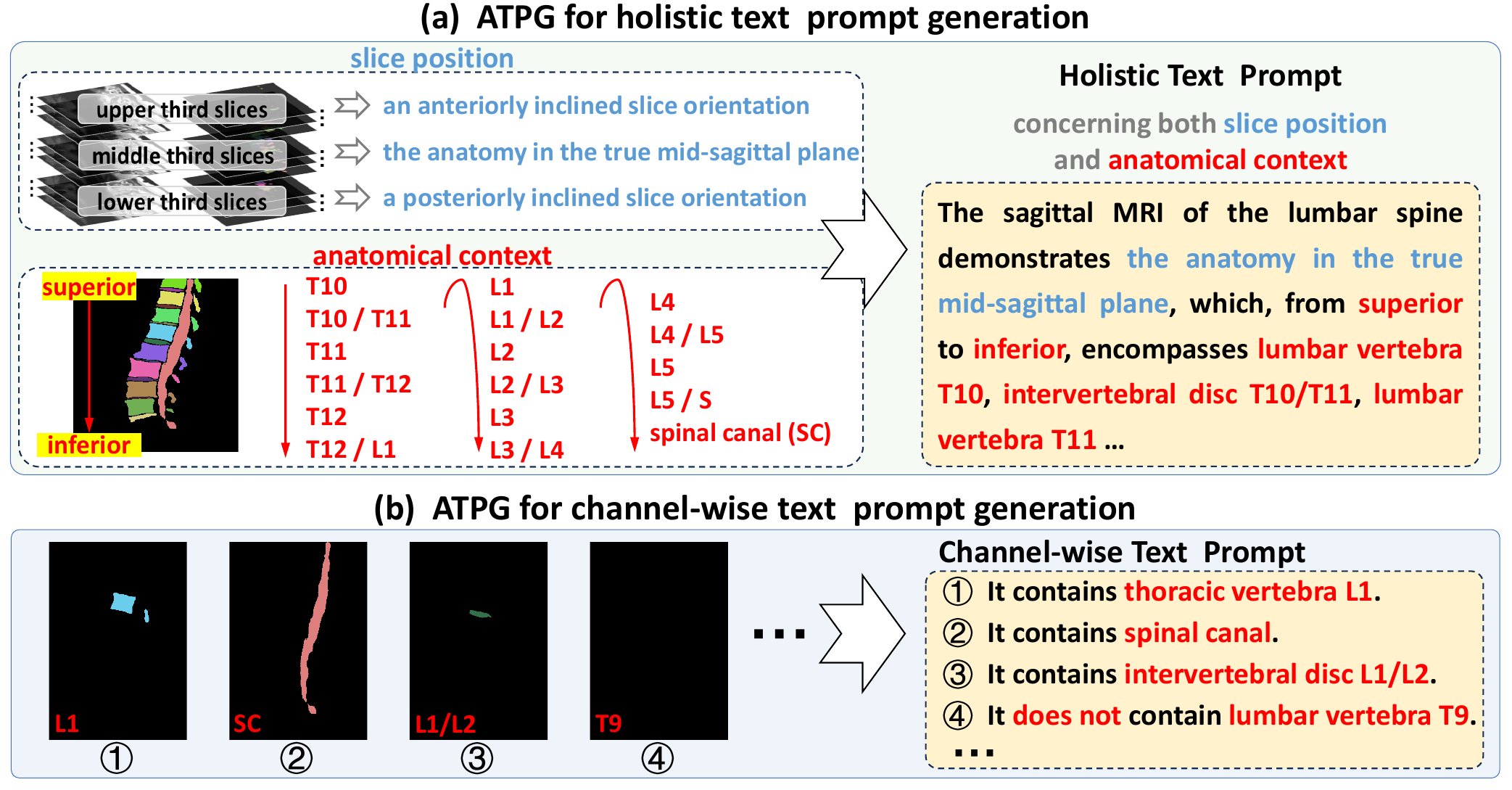}
    \caption{The process of text prompt generation in the ATPG module. }
    \label{fig:prompt-gen}
\end{figure}

ATM-Net features an advanced Anatomy-Aware Text Prompt Generator (ATPG) that automatically and adaptively generates text prompts in holistic (depicted in \Cref{fig:prompt-gen} (a)) and class-wise channel-level views (depicted in \Cref{fig:prompt-gen} (b)). These prompts are carefully aligned with anatomical priors and slice positioning of lumbar spine images. 
% Following the image preprocessing steps detailed in Section~\ref{subsec:preprocess}, ATPG performs this task effectively, as displayed in Figure~\ref{fig:prompt-gen}.

\textbf{ATPG for holistic text prompt generation.}
For the holistic text prompt, ATPG first analyzes the annotation information of the entire image to generate text descriptions on two levels (illustrated in the upper part of \Cref{fig:prompt-gen} (a)). 
The first level establishes spatial perception by determining the approximate position of the slice on the sagittal plane, categorizing it as \textit{upper third slice, middle third slice}, or \textit{lower third slice}.
\textcolor{black}{Crucially, by explicitly encoding such slice-position information, ATPG implicitly injects Z-axis global spatial consistency into the 2D network. 
This design strikes an effective balance between performance and clinical efficiency: it aligns with the sequential slice-by-slice clinical workflow while preserving the computational advantages of 2D networks, yet incorporates essential 3D context to achieve accurate fine-grained segmentation.}
The second level uses the annotation data to provide a top-down description based on the anatomical structure prior, placing the spinal canal at the end of the sequence. Each vertebra and intervertebral discs
is specifically described by type, such as \textit{T11 and L2/L3}, ensuring a systematic and comprehensive description.
% In the lower part of \Cref{fig:prompt-gen} (a), 
An example of a final generated holistic text prompt is as follows.
\begin{quotation} 
\noindent {\color{gray}\textit{``The sagittal MRI of the lumbar spine demonstrates the anatomy in the true mid-sagittal plane, which, from superior to inferior, encompasses lumbar vertebra T10, intervertebral disc T10/T11..."} }
\end{quotation}
These prompts integrate spatial perception and anatomical structure prior knowledge of lumbar spine images, allowing the model to effectively encode essential positional and anatomical information for more precise and robust segmentation.

% \begin{quotation} 
% \noindent \textit{``The sagittal MRI of the lumbar spine demonstrates the anatomy in the true mid-sagittal plane, which, from superior to inferior, encompasses lumbar vertebra T10, intervertebral disc T10/T11..."} \end{quotation}

\textbf{ATPG for channel-wise text prompt generation.}
In this step, ATPG focuses on the annotation information of each individual category within the image. This step provides the model with crucial category presence information by explicitly indicating whether each anatomical structure appears in the image. 
An example of a final generated channel-wise text prompt is as follows.
\begin{quotation}
\vspace{-0.2cm}
   \noindent \textit{\color{gray}{``It contains thoracic vertebra L1."}}
   \vspace{-0.2cm}
\end{quotation}
In this way, ATPG enables class-specific enhancement during contrastive learning, thereby boosting classification accuracy, anatomical structure recognition, and especially the distinction between similar and adjacent structures.
% it enriches both inter-class and intra-class semantic discrimination,
% \Cref{fig:prompt-gen} (b) probides examples of the generated channel-wise text prompt.
% \begin{quotation}
%    \noindent \textit{``It contains thoracic vertebra L1."}
% \end{quotation}

% % \chapquote{``Begin at the beginning,¨ the King said gravely, ``and go on till you come to the end: then stop."}{Lewis Carroll}{Alice in Wonderland}
% \chapquote{``Begin at the beginning,¨ the King said gravely, ``and go on till you come to the end: then stop."}

% By combining these two steps, ATPG generates anatomical text descriptions with prior knowledge and helps the model understand complex lumbar spine images, serving as a crucial support for other ATM-Net modules.
\textcolor{black}{By combining these two steps, ATPG generates anatomical descriptions enriched with prior knowledge, helping the model interpret complex lumbar images and supporting subsequent ATM-Net modules. Unlike simple one-hot encoding, which treats classes as isolated indices and ignores spatial topology, our text prompts capture essential anatomical dependencies. By providing explicit structural priors (e.g., indicating T12 is above L1), ATPG enables the network to effectively capture the continuous spatial structure of the lumbar spine.}
% , leading to better segmentation accuracy.

\textcolor{black}{Furthermore, while ATPG utilizes mask annotations during training, ATM-Net absolutely does not use any ground-truth mask information during inference. In practice, text prompts are generated via two flexible mechanisms: (1) Fully Automatic Mode, which uses a unified, fixed text template with only the raw MRI image as input (identical to pure vision baselines); and (2) Interactive Mode, which constructs prompts via limited clinician intervention (e.g., sparse clicks). This design guarantees both robust clinical applicability and strictly fair performance comparisons.}

\subsection{Holistic Anatomy-Aware Semantic Fusion (HASF)}
\label{subsec:HASF}

In ATM-Net, the encoded visual and text features contain rich semantic and anatomical information from distinct modalities. In this section, we employ the HASF module (Fig.~\ref{fig:framework}(b)) to fuse multi-scale text and visual features, enriching anatomical semantics with ATPG descriptions and LLM knowledge for deeper cross-modal complementarity.
% In this section, we employ the HASF module to combine these multimodal features, achieving a deeper level of information complementary. The design of HASF is illustrated in \Cref{fig:framework}(b).

HASF first aligns the dimensions of the text and visual features.
The LLM encoded text feature $\mathbf{g}_h\in\mathbb{R}^{L\times C}$ is mapped to $\mathbf{f}_{\text{text}}^{H}\in\mathbb{R}^{M\times C_i}$ via \textit{$1*1$ conv, linear transformation}, and \textit{ReLU}, where $M$ tokens have dimension $C_i$ at stage $i$.
For visual features, $\mathbf{f}_v^i$ is mapped from $\mathbb{R}^{H\times W\times C_i}$ to $\mathbb{R}^{(HW)\times C_i}$ and refined by multi-head self-attention $SA(Q,K,V)$:
\begin{equation}
{\mathbf{f}_v^i}'=\mathbf{f}_v^i+Norm(SA(PE(\mathbf{f}_v^i), PE(\mathbf{f}_v^i), \mathbf{f}_v^i)),
\end{equation}
where $Norm(\cdot)$ denotes the normalization layer, and ${\mathbf{f}_v^i}'\in \mathbb{R}^{(H\times W)\times C_i}$ is the visual features enhanced by positional encoding $PE(\cdot)$ and employing multi-scale feature extraction (Eq.\ref{eq:loop}) to mitigate potential information loss.
% encode text insights with various scales of visual features,
% mitigating potential information loss.

Subsequently, HASF uses the multi-head cross-attention mechanism \( CA(Q, K, V) \) to integrate text insights into the enhanced image features, generating \(\mathbf{f}_m^i\):
\begin{equation}
\mathbf{f}_m^i={\mathbf{f}_v^i}'+\alpha(Norm(CA(PE({\mathbf{f}_v^i}'), PE(\mathbf{f}_{text}^{H}),\mathbf{f}_{text}^{H}))),
\end{equation}
where $\alpha$ is a learnable weighting factor. 
Next, the multi-modal feature $\mathbf{f}_m^i\in \mathbb{R}^{(H\times W)\times C_i}$  is reshaped and upsampled to ${\mathbf{f}_m^{i}}'\in \mathbb{R}^{H'\times W'\times C_{i-1}}$ to match the scale of the skip connected feature ${\mathbf{f}_v^{i-1}}$.

% Finally,  ${\mathbf{f}_m^{i}}'$ is concatenated with low-level visual features $\mathbf{f}_v^{i-1}\in \mathbb{R}^{H^{\prime}\times W^{\prime}\times C_{i-1}}$ obtained through skip connections from the visual encoder (depicted in green in Fig.\ref{fig:framework}(b)). The concatenated features are processed through a conv layer $(Conv(\cdot))$ followed by the ReLU activation $(\sigma(\cdot))$ to generate the next stage output $\mathbf{f}_v^{i-1}\in \mathbb{R}^{H^{'}\times W^{'}\times C_{i-1}}$. This process is performed over five stages to encode text insights with various scales of visual features, formulated as:

Finally, ${\mathbf{f}_m^i}'$ is concatenated with skip-connected low-level features $\mathbf{f}_v^{i-1}\!\in\!\mathbb{R}^{H'W'C_{i-1}}$ (depicted in green in Fig.\,\ref{fig:framework}(b)), then refined by $Conv(\cdot)$ and ReLU ($\sigma(\cdot)$) to yield the next stage $\mathbf{f}_v^{i-1}\!\in\!\mathbb{R}^{H'W'C_{i-1}}$. Repeating this across five stages injects text insights into every visual scale, formulated as:
\begin{equation}
% \vspace{-0.3cm}
    \begin{cases}
        \mathbf{f}_v^{i-1} = \sigma(Conv([{\mathbf{f}_m^{i}}', \mathbf{f}_v^{i-1}])) & \text{if } i \in [5,\dots, 1] \\
        \mathbf{f}_o = Softmax(\mathbf{f}_v^i) & \text{if } i = 0,
    \end{cases}
    \label{eq:loop}
    % \vspace{-0.1cm}
\end{equation}

\noindent where $\begin{bmatrix} \cdot, ~\cdot \end{bmatrix}$ denotes the concatenation operation along the channel dimension,
% {\color{cyan}Here, the value of $i \in [5,\dots, 1]$, indicating that we repeatedly apply equations \ref{eq:text-proj} through \ref{eq:loop} to achieve multi-scale text-image feature enhancement. 
% After multiple skip connections, the outputs of the neural network are converted into a probability distribution using the $Softmax(\cdot)$ function.
and the final output \(\mathbf{f}_o\) is obtained by using the $Softmax(\cdot)$ function to the network's output $\mathbf{f}_v^{0}$.
% \begin{equation}
%     \mathbf{f}_o=Softmax(\mathbf{f}_v^{0})
% \end{equation}
% Dice:~\citep{milletari2016v}  Focal~\citep{lin2017focal}
For the loss function, the HASF module combines Dice loss and Focal loss for optimization, which goes as:
\begin{equation}
\label{eq:dice-focal}
\mathcal{L}_{DiceFocal}=\mathcal{L}_{Dice} {(\mathbf{f}_o,\mathbf{y})}+\mathcal{L}_{Focal} {(\mathbf{f}_o,\mathbf{y})}.
\end{equation}

\subsection{Channel-wise Contrastive Anatomy-Aware Enhancement (CCAE)}
\label{subsec:CCAE}
{\color{black}
In fine-grained scenarios, adjacent lumbar substructures (e.g., L4/L5 vertebrae) exhibit high morphological similarity, causing inter-class confusion that holistic text guidance alone cannot fully resolve. 
While HASF integrates global anatomical context, it lacks explicit class-level feature separation. 
We thus propose CCAE to address this through channel-wise contrastive learning, aligning fused image features with class-specific text prompts to maximize inter-class separation and intra-class compactness, with each channel corresponding to one substructure.
}

% In fine-grained scenarios, HASF faces a potential limitation where inconsistencies in multi-modal features may result in the misalignment of specific categories.
% % Thus, CCAE aims to refine the consistency between modalities at the channel level to enhance ATM-Net's discrimination ability and the fine-grained segmentation performance, 
% Thus, we propose CCAE to refine inter-modality consistency at the channel level, thereby enhancing ATM-Net's discriminative power and the precision of fine-grained segmentation (Fig.~\ref{fig:framework} (c)). In ATM-Net's setting, each channel represents one specific substructure.

CCAE shares the same text encoder as HASF. 
To enhance channel consistency and maximize the mutual information between modalities, 
we introduce a multi-modal contrastive loss, aligning $\mathbf{f}_{o}^{'}$ with $\mathbf{f}_{text}^{C}$.
Here, $\mathbf{f}_{o}^{'}$ is the image features incorporated with holistic textual insights, while $\mathbf{f}_{text}^{C}$ is the class-wise channel-level text features.
% representations of image fusion features and text features at the channel level, maximizing their mutual information.
Specifically, the channel-level text features are stacked ($Stack(\cdot)$) along the channel dimension, as formulated in \Cref{eq:fc_text}. 
% to align and reduce the dimensions of both text and visual features. $Norm(\cdot)$ is utilized to standardize the text and visual feature distribution, ensuring the stability of subsequent contrastive loss calculations, and the equations go as follows:
\begin{equation}
\label{eq:fc_text}
\mathbf{f}_{text}^{C}=Norm(GAP(Stack(\mathbf{g}_{p^s}))).
\end{equation}
% Next, we conduct global average pooling ($GAP(\cdot)$) and $Norm(\cdot)$ to align the dimensions and ensure stable contrastive loss calculations, 
Next, we conduct global average pooling ($GAP(\cdot)$) to align and reduce the dimensions of both text and visual features. $Norm(\cdot)$ is utilized to standardize the text and visual feature distribution, ensuring the stability of subsequent contrastive loss calculations, 
as formulated in \Cref{eq:fo}.
\begin{equation}    
\label{eq:fo}
\mathbf{f}_{o}^{'}=Norm\bigl(GAP\bigl(\mathbf{f}_{o}\bigr)\bigr).
\end{equation}
Here, $\mathbf{g}_{p^s}$ are the class-wise channel-level text features, while $\mathbf{f}_{o}$ denote the visual features fused with holistic text insights. 
Next,  $\mathbf{f}_{text}^{C}$ and $\mathbf{f}_{o}^{'}$ are used to compute the class-wise channel-level contrastive loss $L_{ftc}$, formulated as:
% \begin{equation}
% \begin{split}
% \mathcal{L}_{ftc} &= \frac{1}{2s} \sum_{i=1}^{s} \Bigl( \mathcal{L}_{InfoNCE}(\mathbf{f}_{{o}_{i}}^{'}, \mathbf{f}_{text}^{C}) \\
% &\quad + \mathcal{L}_{InfoNCE}({\mathbf{f}_{text}^{C}}_{i}, \mathbf{f}_{o}^{'}) \Bigr),
% \end{split}
% \end{equation}

\begin{equation}
\mathcal{L}_{ftc} = \frac{1}{2s} \sum_{i=1}^{s} \Bigl( \mathcal{L}_{InfoNCE}(\mathbf{f}_{{o}_{i}}^{'}, ~\mathbf{f}_{text}^{C}) + \mathcal{L}_{InfoNCE}({\mathbf{f}_{text}^{C}}_{i}, ~\mathbf{f}_{o}^{'}) \Bigr),
\end{equation}
where $s$ denotes the number of classes.
We enhance InfoNCE loss by adopting class-wise channel-level positive and negative pairs, introducing a multi-modal visual-text contrastive loss. These enhancements improve multi-modal channel consistency and maximize mutual information.

\subsection{Overall Loss Function}
\label{subsec:overloss}
To date, we have introduced two primary training objectives: $\mathcal{L}_{DiceFocal}$ aims to
assess the segmentation performance while mitigating class imbalance issues, whereas $\mathcal{L}_{ftc}$ improves cross-modal feature alignment and consistency. The overall loss function is:
\begin{equation}
    \mathcal{L}_{total}=\lambda_1 * \mathcal{L}_{DiceFocal}+\lambda_2 * \mathcal{L}_{ftc},
\end{equation}
where $\lambda_1=1$ and $\lambda_2=0.2$ in this study. 
% By utilizing both $\mathcal{L}_{DiceFocal}$ and $\mathcal{L}_{ftc}$, we effectively bridge the gap between cross-modal features at both global and local levels, enabling the segmentation model to learn richer semantics and enhance its performance.
In this way, we jointly optimize $\mathcal{L}_{\text{DiceFocal}}$ and $\mathcal{L}_{\text{ftc}}$, aligning the global and local cross-modal insights, and endowing the ATM-Net with richer semantics and higher accuracy.

% {\color{cyan}
% \subsection{Overall loss function}
% To date, we have introduced two primary training objectives for the ATM-Net method:
% \begin{itemize}
%     \item \textit{$\mathcal{L}_{DiceFocal}$:} This serves as the base loss function, aimed at enhancing the accuracy of image segmentation tasks. Specifically, it addresses the issue of class imbalance and also aids in enhancing global cross-modal fusion.
%     \item \textit{$\mathcal{L}_{ftc}$:} This loss function is designed to recalibrate channel-wise features to promote local cross-modal alignment.
% \end{itemize}
% The overall loss function of our method is expressed as:
% \begin{equation}
%     \mathcal{L}_{total}=\lambda_1 * \mathcal{L}_{DiceFocal}+\lambda_2 * \mathcal{L}_{ftc}
% \end{equation}
% where $\lambda_1$ and $\lambda_2$ are hyperparameters that control the contributions of each loss term. In this study, we set $\lambda_1=1$ and $\lambda_2=0.2$. By utilizing both $\mathcal{L}_{DiceFocal}$ and $\mathcal{L}_{ftc}$, we effectively bridge the gap between cross-modal features at both global and local levels. This enables the segmentation model to learn richer semantic information, thereby improving its performance.
% }

\section{Experiment configurations}

\subsection{Datasets}
\label{subsec:dataset}

To assess the segmentation performance and generalization ability of ATM-Net, this study employs two influential lumbar spine datasets (\Cref{tab:dataset_characteristics}), including:
\textit{(1) MRSpineSeg}~\citep{pang2021spineparsenet} consists of 172 MR volumetric images, totaling 2,169 T2-weighted sagittal images.  The dataset comprises 19 categories, consisting of 10 VBs and 9 IDs.  \textit{(2) SPIDER}~\citep{van2024lumbar} consists of 447 MR volumetric images, totaling 12596 T1- and T2-weighted sagittal images.  The dataset includes 19 lumbar spine substructures, comprising 9 VBs, 9 IDs, and 1 SC.

\begin{table}[htbp]
  \centering
  \caption{Datasets characteristics.}
  \label{tab:dataset_characteristics}
  % \resizebox{\linewidth}{!}{
  \begin{tabular}{
    >{\centering\arraybackslash}m{2.0cm}
    >{\centering\arraybackslash}m{1.1cm}
    >{\centering\arraybackslash}m{1.1cm}
    >{\centering\arraybackslash}m{1.9cm}
    >{\centering\arraybackslash}m{1.9cm}
    >{\centering\arraybackslash}m{1.18cm}
  }
    \hline
    \textbf{Dataset} & \textbf{Volumes} & \textbf{Slices} & \textbf{Resolution} & \textbf{Slices / Case} & \textbf{Classes} \\
    \hline
    \textbf{MRSpineSeg} & 172 & 2,169 & $512 \times 512$ -- $1024 \times 1024$ & 12 -- 15 & {19 \footnotemark[1]} \\
    \textbf{SPIDER} & 447 & 12596 & $264 \times 216$ -- $1168 \times 3682$ & 8 -- 154 & 19 \footnotemark[2] \\
    \hline
  \end{tabular}
  \vspace{-0.15cm}
  \footnotetext[1]{{\fontsize{6}{2}\selectfont  MRSpineSeg contains 10 vertebrae, 9 intervertebral discs.}} \vspace{-0.1cm}
  \footnotetext[2]{{\fontsize{6}{2}\selectfont SPIDER contains 9 vertebrae, 9 intervertebral discs, 1 spinal canal}}
  \vspace{-0.2cm}
\end{table}

\begin{figure}[h]
    \centering
    \includegraphics[width=1.00\linewidth,height=2.9cm]{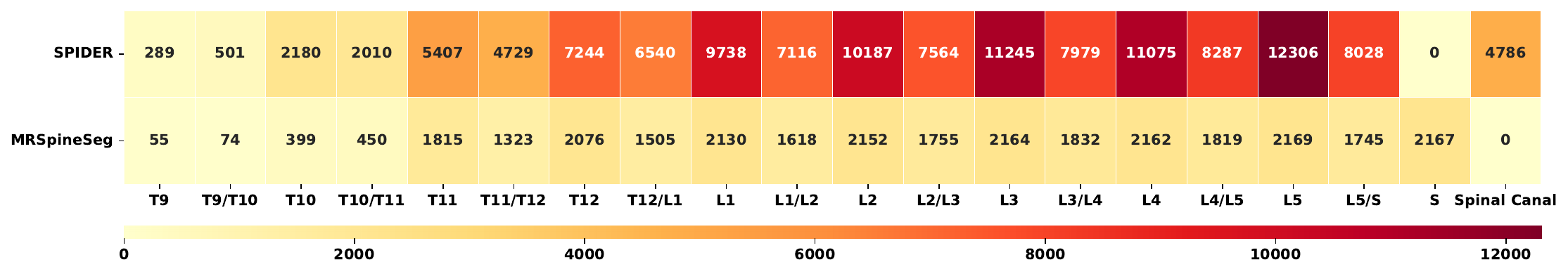}
    \caption{Class distribution of MRSpineSeg and SPIDER. Best viewed in color.}
    \label{fig:class-distribution}
\end{figure}

Notably, not all images across both datasets feature the full 19 classes, and the frequency of different substructures varies significantly among the images (\Cref{fig:class-distribution}).
Additionally, neither dataset includes corresponding text annotations.
ATM-Net applies the ATPG module to automatically and adaptively generate precise descriptions for text prompts. 
These prompts provide information on slice positioning and anatomical prior knowledge, covering both holistic images and fine-grained classes.

% Both datasets feature segmentation annotations but lack corresponding text annotations. Furthermore, not all images across both datasets include all 19 categories, and there is considerable variation in the frequency of different substructures (See \textit{Supplementary}).

% this study employs two influential datasets for evaluation:
% \textit{MRSpineSeg}~\citep{pang2021spineparsenet} and \textit{SPIDER}~\citep{van2024lumbar}. Both datasets focus on the segmentation of lumbar spine MR images. The characteristics of both datasets are summarized in \cref{tab:dataset_characteristics}.

% \textit{\textbf{MRSpineSeg dataset}} consists of 172 MR volumetric images, totaling 2,169 T2-weighted sagittal images and their corresponding annotations. The image sizes range from $512 \times 512$ to $1024 \times 1024$ pixels, with each sequence containing between 12 to 15 slices. The dataset includes 19 classification labels, comprising 10 VBs and 9 IDs. 

\subsection{Image Preprocessing}
\label{subsec:preprocess}

ATM-Net conducts necessary preprocessing steps to ensure segmentation efficacy. 
Following the strategy in \cite{deng2024effective,deng2023modified}, all the slices were cropped and resized to the resolution of $384*384$ before being input. 
We employed a stratified random sampling strategy, dividing the dataset into training, validation, and testing sets in an $8:1:1$ ratio. 
% Such a strategy ensures the consistency of the proportion of samples across categories. 
% could be deleted. Lancer
% Specifically, in the MRSpineSeg dataset, 1729 slices (80\%) were used for training, 207 slices (10\%) for validation, and the remaining 233 slices (10\%) for testing. For the SPIDER dataset, this ratio was maintained, with 10070 slices (80\%) for training, 1250 slices (10\%) for validation, and 1276 slices (10\%) for testing.  

To enhance the model's robustness, we applied random distortions to the data, which introduces slight deformations to mimic the variability in lumbar spine anatomical structures. This approach aids in improving the model’s generalization capability.

% Among the two datasets in this study, SPIDER is more prone to challenging issues such as black borders and irregular noise.
% Traditional methods, which simply crop the edges, struggle to address such uneven noise effectively. We designed a region-based mean cropping algorithm. This algorithm traverses various regions of the image and calculates the average pixel value for each region. If the average pixel value of a region falls below a predefined threshold, that region is identified as a potential black border or invalid area and is cropped accordingly. The pseudo-code is shown in Algorithm~\ref{algo:cropping}, {\color{cyan}where the input is a data dictionary $\mathcal{D}$ containing the image and its ground truth, a block size $\mathbf{B}$, and a threshold percentage $\tau$; the output is the cropped data dictionary $\mathcal{D'}$ with the image and ground truth cropped based on the local mean calculation.}
% The core advantage of this method is its ability to effectively filter out randomly distributed highlight noise.

\subsection{Implementation details}

We implement ATM-Net and corresponding comparing experiments with several open-source libraries, including but not limited to PyTorch (version 2.1.0)\footnote{PyTorch: https://pytorch.org/} and MONAI (version 1.3.0)\footnote{MONAI: https://monai.io/}. 
PyTorch Lightning (version 2.1.2)\footnote{PyTorch Lightning: https://lightning.ai/docs/pytorch/stable} was employed as the final training and inference framework to streamline the development process. All methods were trained on a device with Hygon C86 7360 24-core processor, and four NVIDIA GeForce RTX 3090 GPUs, each with 24GB of memory. 
% {\color{blue}Additionally, the system utilizes the Hygon C86 7360 24-core Processor, which features a total of 96 threads.}

For model training, we combined Dice loss and Focal loss as our loss function to balance the classification between foreground and background. We used the AdamW optimizer with a batch size of 8 and an initial learning rate of 1e-4, which was gradually decreased to 1e-6 using a cosine annealing strategy to facilitate model convergence.

\subsection{Evaluation metrics}
% To demonstrate the effectiveness of ATM-Net, we utilize four evaluation metrics from two perspectives.
% The first focuses on \textit{region overlap}, including the Dice Similarity Coefficient (DSC) and Jaccard Index (Jaccard). Both measure the intersection area between the predicted mask and the ground truth (GT) region, with the DSC being sensitive to the segmentation performance of small objects. 
% The other is from the perspective of \textit{boundary similarity}, including $95\%$ Hausdorff Distance (HD95), and Average Surface Distance (ASD). Both measure the accuracy and consistency of the segmentation boundaries.

To demonstrate the effectiveness of ATM-Net, we employ metrics from two perspectives. The first focuses on \textit{\textbf{region overlapping}}, including the Dice Similarity Coefficient (DSC) and Jaccard Index (Jaccard). These metrics measure the intersection area between the predicted mask and the ground truth (GT) region, with the DSC being more sensitive to the segmentation performance of small targets. The second assesses \textit{\textbf{boundary similarity}}, including the $95\%$ Hausdorff Distance (HD95) and Average Surface Distance (ASD) being adopted to evaluate the accuracy and consistency of the segmentation boundaries.

\section{Experimental results and analysis}

\subsection{Comparing experiments}
\label{subsec:compare-exp}

% To evaluate the segmentation performance of ATM-Net, we compare it with two types of methods: (1) models that have been long validated for effectiveness in medical image segmentation tasks, including U-Net~\cite{ronneberger2015u}, UNETR~\cite{hatamizadeh2022unetr}, SegResNet~\cite{myronenko20193d}, Attention U-Net~\cite{oktay2018attention}, Swin UNETR~\cite{tang2022self}, nnU-NetV2~\cite{isensee2024nnu}, 
% and (2) solutions specifically proposed for lumbar spine substructure segmentation.

To evaluate the segmentation performance of ATM-Net, we conduct the comparison experiments with 10 promising solutions, categorized into three types for clarity:
(1) \textit{General medical image segmentation (MIS) methods (marked as $\bullet$):} The first six models serve as fundamental references, which have long been validated for effectiveness in medical image segmentation tasks. 
(2) \textit{Fine-grained spine segmentation-specific methods (marked as \scalebox{0.8}{$\blacktriangle$}):} Note that methods specifically for this task are limited, and due to reasons such as non-open-sourced code and insufficient algorithm details, we included three methods specifically for lumbar spine substructure segmentation.
{\color{black}
(3) \textit{VLM/VFM-based methods (marked as $\star$):} Since no visual-language model (VLM) or visual foundation model (VFM) has been specifically designed for fine-grained lumbar spine segmentation to date, we adopted three representative models as additional baselines. These include Ariadne's Thread (a VLM originally developed for segmenting infected regions in chest X-ray images), TeViA (a VLM that efficiently integrates textual insights through a segmentation-specific text-to-vision alignment mechanism), and H-SAM (a SAM-based VFM that achieves prompt-free medical image segmentation via a two-stage hierarchical decoding mechanism). Notably, to ensure a strictly fair comparison, both Ariadne's Thread and TeViA were evaluated using the exact same Option 3 text prompts (see \Cref{tab:prompt-opt}) as our ATM-Net.
% Since no visual-language model (VLM) or visual foundation model (VFM) has been specifically designed for fine-grained lumbar spine segmentation to date, we adapt two representative models as additional baselines, including Ariadne's Thread, a VLM originally developed for infection region segmentation in chest X-ray images, and IMIS-Net, a VFM based on the SAM architecture that generates high-quality segmentation results by leveraging multiple interactive cues.
}
Brief descriptions of all the 11 methods are provided below:

\begin{itemize}
    \item \textit{U-Net~\citep{ronneberger2015u}:} Widely used for biomedical image segmentation, combining a contracting and expanding path for context and precise localization.
    \item \textit{UNETR~\citep{hatamizadeh2022unetr}:} Transformer-based model for volumetric MIS.
    \item \textit{SegResNet~\citep{myronenko20193d}:} Extension of ResNet for improved feature representation in MIS.
    \item \textit{Attention U-Net~\citep{oktay2018attention}:} Enhanced U-Net incorporating attention gates for better context capture and refined segmentation.
    \item \textit{Swin UNETR~\citep{tang2022self}:} Combines Swin Transformer with UNETR for state-of-the-art performance.
    \item \textit{nnU-NetV2~\citep{isensee2024nnu}:} Widely adopted for its robustness and adaptability in MIS.
    \item[\scalebox{0.7}{$\blacktriangle$}] \textit{Modified BiSeNet~\citep{deng2023modified}:} Modified BiSeNet for multi-class segmentation of vertebrae and intervertebral discs in MRI.
    \item[\scalebox{0.7}{$\blacktriangle$}] \textit{U-BiSeNet~\citep{deng2024effective}:} A network combining U-Net and BiSeNet is proposed for multi-class segmentation of vertebrae and intervertebral discs in MRI.
    \item[\scalebox{0.7}{$\blacktriangle$}] \textit{SpineParseNet~\citep{pang2021spineparsenet}:} A two-stage framework that enhances VBs and IDs segmentation in MRI through 3D coarse and 2D refined processing.
    % \textcolor{red}{
    % \item[\scalebox{0.9}{$\star$}] \textit{IMIS-Net~\citep{cheng2025interactive}:} A segmentation model based on the SAM architecture that generates high-quality segmentation results by leveraging multiple interactive cues.
    % }
    \item[\scalebox{0.9}{$\star$}] \textit{Ariadne’s Thread~\citep{zhong2023ariadne}:} A language-driven multimodal approach with multi-level fusion to enhance infection region segmentation in chest X-ray images.
    \textcolor{black}{
    \item[\scalebox{0.9}{$\star$}] \textit{TeViA~\citep{zeng2025harnessing}:} By introducing a segmentation-specific text-to-vision alignment mechanism and momentum-updated visual prototypes, it efficiently integrates the semantic insights of text prompts with visual features.    
    \item[\scalebox{0.9}{$\star$}] \textit{H-SAM~\citep{cheng2024unleashing}:} Efficiently adapting SAM via a two-stage hierarchical decoding mechanism to achieve high-performance medical image segmentation in a prompt-free manner.
    }
    
\end{itemize}

\begin{table*}[htbp]
  \centering
  \caption{{\color{black}Quantitative comparison of ATM-Net and comparative methods (performance, complexity). 
  Symbols: \scalebox{0.86}{$\bullet$} = General MIS solutions, \scalebox{0.7}{$\blacktriangle$} = Lumbar-specific, \scalebox{0.9}{$\star$} = VLM/VFM. 
  Best/second-best: \textbf{bold}/\underline{underlined}.}}
  \label{tab:comprehensive-compare}
  \footnotesize
  \renewcommand\arraystretch{1.2}
  \setlength{\tabcolsep}{3pt} % 进一步缩小列间距
  \begin{tabular}{
    l | rrrr | rrrr | rrr
  }
    \toprule
    \multirow{2}{*}{\textbf{Method}} &
    \multicolumn{4}{c|}{\textbf{MRSpineSeg}} & \multicolumn{4}{c|}{\textbf{SPIDER}} & \multicolumn{3}{c}{\textbf{Complexity}} \\
    \cmidrule(lr){2-9} \cmidrule(lr){10-12}
    & \makecell{DSC\\{\tiny{\%$\uparrow$}}} & \makecell{Jacc\\{\tiny{\%$\uparrow$}}} & \makecell{HD95\\{\tiny{px$\downarrow$}}} & \makecell{ASD\\{\tiny{px$\downarrow$}}}
    & \makecell{DSC\\{\tiny{\%$\uparrow$}}} & \makecell{Jacc\\{\tiny{\%$\uparrow$}}} & \makecell{HD95\\{\tiny{px$\downarrow$}}} & \makecell{ASD\\{\tiny{px$\downarrow$}}}
    & \makecell{Para\\{\tiny{(M)}}} & \makecell{FLOPs\\{\tiny{(G)}}} & \makecell{Time\\{\tiny{(ms)}}} \\
    \midrule
    % 通用MIS模型
    U-Net \cite{ronneberger2015u} \scalebox{0.8}{$\bullet$} & 58.59 & 47.76 & 32.65 & 8.35 & 52.52 & 42.23 & 49.94 & 25.04 & 1.49 & 37.09 & 8.93 \\
    UNETR \cite{hatamizadeh2022unetr} \scalebox{0.8}{$\bullet$} & 61.81 & 53.80 & 24.38 & 8.25 & 60.16 & 52.05 & 15.49 & 3.92 & 87.12 & 473.47 & 42.5 \\
    SegResNet \cite{myronenko20193d} \scalebox{0.8}{$\bullet$} & 63.18 & 55.25 & 14.66 & 4.68 & 61.58 & 54.04 & 13.62 & 4.16 & 0.39 & 17.90 & 16.24 \\
    Att. U-Net \cite{oktay2018attention} \scalebox{0.8}{$\bullet$} & 63.73 & 55.08 & 40.84 & 20.96 & 62.72 & 54.81 & 15.67 & 4.44 & 4.46 & 145.25 & 19.73 \\
    Swin UNETR \cite{tang2022self} \scalebox{0.8}{$\bullet$} & 64.58 & 56.78 & 17.32 & 7.04 & 66.67 & 59.31 & 10.29 & 3.21 & 25.12 & 340.7 & 65.87 \\
    nnUNetV2 \cite{isensee2024nnu} \scalebox{0.8}{$\bullet$} & 76.43 & 67.50 & 16.70 & 3.91 & 71.59 & 63.49 & 14.28 & 4.52 & 19.10 & 412.7 & 85.83 \\
    % \midrule
    % 腰椎专用模型
    Mod-BiSeNet \cite{deng2023modified} \scalebox{0.6}{$\blacktriangle$} & 65.49 & 57.11 & 15.36 & 5.10 & 64.56 & 56.03 & 12.24 & 3.40 & 41.91 & 273.95 & 30.31 \\
    U-BiSeNet \citep{deng2024effective} \scalebox{0.6}{$\blacktriangle$} & 66.03 & 57.67 & 20.20 & 9.45 & 64.67 & 56.20 & \underline{10.19} & 3.08 & 12.59 & 186.87 & 16.11 \\
    SpineParseNet \cite{pang2021spineparsenet} \scalebox{0.6}{$\blacktriangle$} & \underline{78.84} & \underline{71.65} & 17.54 & 6.10 & 71.08 & 63.16 & 14.05 & 3.15 & 15.95 & 392.85 & 312.48 \\
    % \midrule
    % VLM模型
    A-Thread \cite{zhong2023ariadne} \scalebox{0.9}{$\star$} & 70.56 & 62.90 & \underline{9.77} & 2.52 & 62.82 & 56.42 & 10.25 & 3.18 & 115.39 & 634.20 & 54.51 \\
    \textcolor{black}{TeViA} \citep{zeng2025harnessing} \scalebox{0.8}{$\star$} & 72.44 & 63.72 & 10.07 & \underline{2.35} & \underline{77.74} & \underline{68.75} & 10.68 & 3.15 & 114.48 & 79.15 & 44.74 \\
    \textcolor{black}{H-SAM} \citep{cheng2024unleashing} \scalebox{0.8}{$\star$} & 70.51 & 62.75 & 14.19 & 3.53 & 64.01 & 57.36 & 10.43 & \underline{2.84} & 116.18 & 105.93 & 270.20 \\
    % \midrule
    % 本文方法
    \rowcolor{gray!20} ATM-Net (Ours) & $\mathbf{81.72}$ & $\mathbf{72.25}$ & $\mathbf{9.60}$ & $\mathbf{2.15}$ & $\mathbf{79.39}$ & $\mathbf{70.56}$ & $\mathbf{9.91}$ & $\mathbf{2.77}$ & 64.87 & 623.95 & 75.00 \\
    \bottomrule
  \end{tabular}
  \vspace{1mm}
  \scriptsize{DSC=Dice Similarity Coefficient; Jacc=Jaccard index; HD95=95th percentile Hausdorff Distance; ASD=Average Surface Distance; Para=Parameters; Time=Inference time.}
\end{table*}

\begin{table*}[p]
\centering
\begin{minipage}[b]{0.20\textwidth}
\centering
\rotatebox{90}{%
%%%%% 第一张
\begin{minipage}{1.05\textheight}
\centering
\caption{{\small{DSC (\%) comparisons for all substructures in MRSpineSeg. The best results are highlighted in \textbf{bold.}}}}
\vspace{-0.13cm}
\label{tab:sub-mr-dsc}
\resizebox{1.02\textheight}{!}{%
\renewcommand{\arraystretch}{0.82} 
\begin{tabular}{c|cccccccccccccccccccc}
\hline Method & S & L5 & L4 & L3 & L2 & L1 & T12 & T11 & T10 & T9 & L5/S & L4/L5 & L3/L4 & L2/L3 & L1/L2 & T12/L1 & T11/T12 & T10/T11 & T9/T10 & Avg. \\
\hline U-Net & 82.31 & 75.30 & 60.96 & 53.87 & 51.36 & 53.20 & 57.21 & 63.43 & 40.53 & 18.3 & 80.00 & 76.97 & 73.34 & 67.43 & 66.98 & 69.81 & 64.73 & 57.30 & 0.19 & 58.59 \\
UNETR & 80.68 & 72.14 & 64.80 & 64.72 & 62.08 & 61.21 & 65.02 & 71.54 & 0 & 53.69 & 74.43 & 71.08 & 73.36 & 72.61 & 72.47 & 72.58 & 74.88 & 67.07 & 0 & 61.81 \\
SegResNet & 82.89 & 83.86 & 78.05 & 75.16 & 69.47 & 65.11 & 61.71 & 69.13 & 0 & 0 & 84.31 & 83.22 & \textbf{86.83} & 86.83 & 78.11 & 71.43 & 73.48 & 50.78 & 0 & 63.18 \\
Attention U-Net & 85.48 & 82.94 & 77.47 & 74.12 & 70.67 & 71.21 & 64.49 & 70.01 & 38.06 & 6.95 & 84.45 & 83.41 & 86.15 & 84.31 & 85.14 & 76.60 & 69.41 & 0 & 0 & 63.73 \\
Swin UNETR & 84.98 & 78.81 & 70.89 & 68.08 & 65.52 & 64.48 & 68.74 & 74.78 & 49.35 & 0 & 79.84 & 76.95 & 78.16 & 75.05 & 73.78 & 73.46 & 75.50 & 68.23 & 0 & 64.58 \\
nnU-NetV2 & 87.58 & 85.81 & 80.45 & 81.29 & 80.86 & 79.44 & \textbf{81.63} & 80.04 & 65.25 & 59.39 & 83.32 & 82.67 & 86.47 & 84.76 & 85.90 & 85.24 & 84.09 & 77.93 & 0 & 76.43 \\ \hline
 Modified BiSeNet & 86.43     & 84.47     & 79.83     & 79.73     & 72.78     & 64.62     & 66.21     & 69.63     & 30.13     & 0         & 83.93     & 84.04     & 86.13     & 83.93     & 75.61     & 68.92     & 72.22     & 55.76     & 0         & 65.49 \\
     U-BiSeNet & 86.13     & 84.35     & 79.32     & 78.18     & 70.84     & 65.23     & 66.82     & 69.65     & 32.99     & 0         & 84.77     & 84.24     & 86.69     & 86.55     & 78.69     & 69.77     & 72.2      & 57.29     & 0.84      & 66.03 \\
     SpineParseNet & \textbf{88.96} & 85.92 & \textbf{83.34} & \textbf{82.46} & 80.21     & \textbf{83.09} & 76.48     & 77.64     & \textbf{78.78} & 31.73     & 81.94     & 82.95     & 83.73     & 83.8      & 85.29     & 85.02     & \textbf{87.23} & 84.37     & 55.09     & 78.84 \\ \hline
    Ariadne’s Thread & 86.96 & \textbf{85.98} & 82.69 & 81.5  & \textbf{81.34} & 82.3  & 80.85 & 78.43 & 0     & 0     & \textbf{93.82} & 84.84 & 86.67 & 88.17 & 87.74 & 86.06 & 85.96 & 79.33 & 0     & 70.56 \\ 
    \textcolor{black}{TeViA} & 86.96 & 84.75 & 81.04 & 79.81 & 78.61 & 76.03 & 74.93 & 76.76 & 48.91 & 20.63 & 84.45 & 84.06 & 86.55 & 87.57 & 85.53 & 83.91 & 80.91 & 67.88 & 0.07  & 72.44 \\ 
    \textcolor{black}{H-SAM} & 83.78 & 83.97 & 81.01 & 79.53 & 76.01 & 76.79 & 79.27 & 77.94 & 72.61 & 0     & 83.58 & \textbf{85.57} & 86.55 & 85.16 & 87.78 & 86.91 & 85.03 & 0     & 0     & 70.51 \\ \hline
     ATM-Net (Ours)      & 87.18     & 85.08     & 81.64     & 82.1      & 80.29     & 76.26     & 79.25     & \textbf{81.04} & 67.03     & \textbf{80.18} & 85.01 & 84.45 & 85.25     & \textbf{89.13} & \textbf{87.96} & \textbf{88.11} & 86.14     & \textbf{85.55} & \textbf{61.02} & \textbf{81.72} \\
% Ours & 87.18 & 85.08 & 81.64 & 82.1 & 80.29 & 76.26 & 79.25 & 81.04 & 67.03 & 80.18 & 85.01 & 84.45 & 85.25 & 89.13 & 87.96 & 88.11 & 86.14 & 85.55 & 61.02 & 81.72 \\
\hline
\end{tabular}
}
\end{minipage}%
}
\end{minipage}%
\hfill
%%%%% 第二张
\begin{minipage}[b]{0.20\textwidth}
\centering
\rotatebox{90}{%
\begin{minipage}{1.05\textheight}
\centering
\caption{{\small{Jaccard (\%) comparisons for all substructures in MRSpineSeg. The best results are highlighted in \textbf{bold}.}}}
\vspace{-0.13cm}
\label{tab:sub-mr-jcc}
\resizebox{1.02\textheight}{!}{%
\renewcommand{\arraystretch}{0.82} 
\begin{tabular}{c|cccccccccccccccccccc}
\hline Method & S & L5 & L4 & L3 & L2 & L1 & T12 & T11 & T10 & T9 & L5/S & L4/L5 & L3/L4 & L2/L3 & L1/L2 & T12/L1 & T11/T12 & T10/T11 & T9/T10 & Avg. \\
\hline U-Net & 70.55 & 62.48 & 48.08 & 40.83 & 38.36 & 40.73 & 45.32 & 52.41 & 29.84 & 10.46 & 68.84 & 64.6  & 61.75 & 55.7  & 56.22 & 58.9  & 53.96 & 48.36 & 0.1   & 47.76 \\
UNETR & 69.42 & 62.12 & 56.21 & 56.76 & 54.15 & 53.34 & 56.52 & 61.68 & 0     & 39.25 & 64.71 & 62.52 & 65.98 & 65.4  & 65.19 & 64.93 & 65.77 & 58.22 & 0     & 53.8 \\
SegResNet & 71.59 & 73.5  & 67.84 & 65.06 & 58.77 & 55.39 & 53.11 & 59.66 & 0     & 0     & 74.05 & 72.85 & 78.07 & 78.33 & 69.47 & 63.89 & 64.73 & 43.51 & 0     & 55.25 \\
Attention U-Net & 75.26 & 72.32 & 66.69 & 63.61 & 59.67 & 59.68 & 54.27 & 59.53 & 28.93 & 3.76  & 74.02 & 72.7  & 76.92 & 75    & 76.22 & 67.28 & 60.71 & 0     & 0     & 55.08 \\
Swin UNETR & 74.67 & 69.39 & 61.74 & 59.56 & 57.22 & 55.85 & 60.33 & 64.63 & 39.58 & 0     & 70.47 & 67.96 & 70.63 & 68.06 & 66.57 & 65.97 & 66.78 & 59.46 & 0     & 56.78 \\
nnU-NetV2 & 78.47 & 76.31 & 71.31 & 72.57 & 71.96 & 70.75 & \textbf{72.5} & 70.87 & 54.41 & 43.9  & 73.61 & 72.93 & 77.86 & 77.06 & 78.13 & 76.97 & 74.47 & 68.37 & 0     & 67.5 \\ \hline
 Modified BiSeNet & 76.58 & 73.99 & 69.56 & 69.44 & 61.78 & 55.52 & 57.28 & 60    & 23.29 & 0     & 73.32 & 73.48 & 76.82 & 74.72 & 67.34 & 61.19 & 63.17 & 47.65 & 0     & 57.11 \\
    U-BiSeNet & 76.19 & 73.88 & 68.83 & 68.04 & 61.1  & 56.09 & 57.71 & 60.21 & 25.62 & 0     & 74.34 & 73.8  & 77.19 & 77.81 & 70.49 & 61.98 & 62.98 & 49.04 & 0.44  & 57.67 \\
    SpineParseNet & \textbf{80.71} & \textbf{78.35} & \textbf{75.97} & \textbf{75.13} & \textbf{72.78} & \textbf{75.8} & 69.49 & 70.05 & \textbf{68.18} & 26.84 & 74.25 & \textbf{75.96} & 77.64 & 77.91 & 79.29 & \textbf{79.24} & \textbf{80.78} & \textbf{76.09} & 46.88 & 71.65 \\ \hline
Ariadne’s Thread & 77.73 & 76.28 & 73.45 & 72.6  & 72.3  & 70.98 & 71.95 & 70.29 & 0     & 0     & 74.05 & 75.28 & \textbf{78.15} & 80.22 & 79.73 & 77.5  & 76.18 & 68.37 & 0     & 62.9 \\ 
\textcolor{black}{TeViA} & 77.47 & 74.99 & 71.49 & 70.53 & 69.29 & 66.11 & 65.37 & 67.01 & 40.16 & 15.28 & 74.57 & 74.06 & 78.01 & 79.33 & 77.03 & 75.43 & 71.09 & 58.17 & 0.05  & 63.72 \\ 
\textcolor{black}{H-SAM} & 72.8  & 73.69 & 71.97 & 70.42 & 67.37 & 68.31 & 69.7  & 67.5  & 60    & 0     & 73.12 & 75.75 & 77.53 & 76.23 & 79.36 & 78.11 & 75.77 & 0     & 0     & 62.75 \\ \hline
     ATM-Net (Ours)      & 77.85 & 75.39 & 72.1  & 72.65 & 70.85 & 66.86 & 69.84 & \textbf{71.04} & 57.1  & \textbf{67.28} & \textbf{75} & 74.51 & 76.91 & \textbf{81.03} & \textbf{79.83} & \textbf{79.24} & 76.43 & 75.58 & \textbf{53.23} & \textbf{72.25} \\
% Ours & 87.18 & 85.08 & 81.64 & 82.1 & 80.29 & 76.26 & 79.25 & 81.04 & 67.03 & 80.18 & 85.01 & 84.45 & 85.25 & 89.13 & 87.96 & 88.11 & 86.14 & 85.55 & 61.02 & 81.72 \\
\hline
\end{tabular}
}
\end{minipage}%
}
\end{minipage}%
\hfill
%%%%% 第三张
\begin{minipage}[b]{0.20\textwidth}
\centering
\rotatebox{90}{%
\begin{minipage}{1.05\textheight}
\centering
\caption{{\small{HD95 (mm) for comparisons all substructures in MRSpineSeg. The best results are highlighted in \textbf{bold}.}}}
\vspace{-0.13cm}
\label{tab:sub-mr-hd95}
\resizebox{1.02\textheight}{!}{%
\renewcommand{\arraystretch}{0.82} 
\begin{tabular}{c|cccccccccccccccccccc}
\hline Method & S & L5 & L4 & L3 & L2 & L1 & T12 & T11 & T10 & T9 & L5/S & L4/L5 & L3/L4 & L2/L3 & L1/L2 & T12/L1 & T11/T12 & T10/T11 & T9/T10 & Avg. \\
\hline U-Net & 47.48 & 35.56 & 42.13 & 49.88 & 51.02 & 49.51 & 52.32 & 39.39 & 40.03 & 57.17 & 7.67  & 10.91 & 15.51 & 17.47 & 17.25 & 14.05 & 13.98 & 13.14 & 45.94 & 32.65 \\
UNETR & 27.35 & 22.43 & 26.08 & 25.77 & 29.2  & 22.31 & 24.86 & 15.01 & {NaN} & 131.87 & 10.3  & 12.59 & 11.82 & 11.81 & 11.19 & 9.97  & 7.86  & 14.09 & {NaN} & 24.38 \\
SegResNet & 21.55 & 16.97 & 20.45 & 24.5  & 25.51 & 24.55 & 20.79 & 16.14 & {NaN} & {NaN} & 5.26  & 6.54  & 5.15  & 5.32  & 7.57  & 9.89  & 7.87  & 16.49 &{NaN} & 14.66 \\
Attention U-Net & 20.95 & 22.03 & 22.8  & 26.68 & 30.49 & 31.07 & 22.04 & 18.01 & 38    & 46.55 & 5.33  & 6.28  & 6.41  & 6.8   & 5.29  & 8.43  & 9.86  & {NaN} & 408.2 & 40.84 \\
Swin UNETR & 17.28 & 18.1  & 25.36 & 25.82 & 25.18 & 21.12 & 19.12 & 17.34 & 50.57 & {NaN} & 8.14  & 9.68  & 9.53  & 10.7  & 10.42 & 9.66  & 7.62  & 8.84  & {NaN} & 17.32 \\
nnU-NetV2 & 22.28 & 22.64 & 26.18 & 24.77 & 21.59 & 17.2  & 17.26 & 15.78 & 23.44 & 38.34 & 10.97 & 9.64  & 8.4   & 6.3   & 5.3   & 5.79  & 8.88  & 15.92 & {NaN} & 16.7 \\ \hline
 Modified BiSeNet & 18.77 & 13.91 & 17.54 & 17.44 & 22.17 & 19.29 & 18.53 & 14.61 & 40.57 & {NaN} & 5.09  & 5.82  & 5.6   & 5.75  & 8.94  & 11.92 & 8.29  & 26.83 & {NaN} & 15.36 \\
    U-BiSeNet & 26.26 & 16.67 & 20.64 & 19.42 & 21.07 & 18.8  & 17.15 & 16.75 & 39.11 & {NaN} & \textbf{4.66}  & \textbf{5.42} & 5.17  & 4.78  & 7.51  & 11.41 & 8.31  & 12.6  & 107.9 & 20.2 \\
    SpineParseNet & 25.43 & 22.5  & 27.02 & 26.4  & 27.25 & 19.5  & 20.67 & 18.59 & 22.01 & 31.89 & 11    & 12.01 & 13.85 & 10.62 & 9.61  & 15.78 & 4.31  & 6.49  & 8.4   & 17.54 \\ \hline
    Ariadne’s Thread & 16.24 & 14.28 & 16.33 & 16.15 & 14.62 & \textbf{12.24} & \textbf{10.65} & \textbf{9.98}  & {NaN}   & {NaN}   & 6.12  & 6.93  & 5.92  & 5.34  & 4.9   & 4.13  & 4.13  & 7.42  & {NaN}   & 9.77 \\ 
    \textcolor{black}{TeViA} & 15.33 & 14.8  & \textbf{13.72} & 16.85 & \textbf{14.31} & 14.59 & 13.41 & 11.3  & 16.41 & 15.84 & 5.68  & 5.57  & 5.05  & 4.39  & 4.36  & 4.16  & 4.14  & 4.15  & 6.32  & 10.07 \\ 
    \textcolor{black}{H-SAM} & 30.85 & 25.29 & 25.76 & 23.92 & 26.67 & 19.47 & 18.78 & 19.25 & 23.89 & NaN   & 7.94  & 7.08  & 7.01  & 13.93 & 5.76  & 5.16  & 5.88  & NaN   & NaN   & 14.19 \\ \hline
     ATM-Net (Ours)      & \textbf{14.19} & \textbf{13.75} & 16.93 & \textbf{16.04} & 16.06 & 14.27 & 11.42 & 10.78 & \textbf{15.08} & \textbf{15.44} & 5.03 & 5.73  & \textbf{4.54} & \textbf{3.85} & \textbf{3.63} & \textbf{3.06} & \textbf{3.1} & \textbf{3.95} & \textbf{5.65} & \textbf{9.6} \\
% Ours & 87.18 & 85.08 & 81.64 & 82.1 & 80.29 & 76.26 & 79.25 & 81.04 & 67.03 & 80.18 & 85.01 & 84.45 & 85.25 & 89.13 & 87.96 & 88.11 & 86.14 & 85.55 & 61.02 & 81.72 \\
\hline
\end{tabular}
}
\end{minipage}%
}
\end{minipage}%
\hfill 
%%%%% 第四张
\begin{minipage}[b]{0.20\textwidth}
\centering
\rotatebox{90}{%
\begin{minipage}{1.05\textheight}
\caption{{\small{ASD (mm) comparisons for all substructures in MRSpineSeg. The best results are highlighted in \textbf{bold}.}}}
\vspace{-0.13cm}
\centering
\label{tab:sub-mr-asd}
\resizebox{1.02\textheight}{!}{%
\renewcommand{\arraystretch}{0.82} 
\begin{tabular}{c|cccccccccccccccccccc}
\hline Method & S & L5 & L4 & L3 & L2 & L1 & T12 & T11 & T10 & T9 & L5/S & L4/L5 & L3/L4 & L2/L3 & L1/L2 & T12/L1 & T11/T12 & T10/T11 & T9/T10 & Avg. \\
\hline U-Net & 8.27  & 8.11  & 11.89 & 14.89 & 17.45 & 18.24 & 16.52 & 11.73 & 8.67  & \textbf{1.18}  & 1.34  & 1.43  & 4.27  & 4.44  & 6.01  & 3.53  & 6.19  & 4.89  & 9.6   & 8.35 \\
UNETR & 5.48  & 6.41  & 9.07  & 9.78  & 11.34 & 9.67  & 10.03 & 4.74  & {NaN} & 19.92 & 4.13  & 6.84  & 7.36  & 7.67  & 7.71  & 6.77  & 4.47  & 8.9   & {NaN} & 8.25 \\
SegResNet & 4.21  & 3.74  & 4.72  & 5.88  & 7.04  & 7.49  & 7.32  & 5.73  & {NaN} & {NaN} & 1.04  & 1.24  & 1.35  & 2.07  & 2.67  & 6.26  & 5.08  & 8.95  & {NaN} & 4.68 \\
Attention U-Net & 4.14  & 3.87  & 5.33  & 6.87  & 8.9   & 10.02 & 6.62  & 4.97  & 5.07  & 3.16  & 1.05  & 1.27  & 1.82  & 1.8   & 1.17  & 2.59  & 5.24  & {NaN} & 303.4 & 20.96 \\
Swin UNETR & 3.04  & 4.64  & 8.32  & 8.73  & 10.2  & 8.47  & 7.41  & 5.16  & 19.57 & {NaN} & 3.38  & 5.07  & 5.57  & 6.62  & 7.21  & 6.75  & 4.68  & 4.91  & {NaN} & 7.04 \\
nnU-NetV2 & 3.91  & 4.19  & 5.88  & 5.96  & 5.65  & 3.65  & 4.13  & 4.42  & 6.96  & 2.25  & 3.42  & 2.85  & 2.72  & 1.97  & 1.84  & 1.77  & 2.97  & 5.77  & {NaN} & 3.91 \\ \hline
Modified BiSeNet & 3.49  & 2.57  & 3.86  & 4.09  & 5.72  & 7.43  & 6.88  & 4.88  & 4.85  & {NaN} & 1.11  & 1.23  & 1.21  & 1.43  & 4.98  & 8     & 5.13  & 19.74 & {NaN} & 5.1 \\
U-BiSeNet & 4.43  & 2.75  & \textbf{1.07} & 5.14  & 7.25  & 6.74  & 6.48  & 5.44  & 5.87  & {NaN} & 1.09  & 1.28  & 1.07  & 1.38  & 3.56  & 7.03  & 5.16  & 6.68  & 94.34 & 9.45 \\
SpineParseNet & 3.87  & 6.52  & 7.68  & 7.39  & 8.48  & 6.96  & 8.33  & 6.52  & 6.85  & 5.03  & 5.42  & 5.87  & 6.89  & 6.3   & 6.14  & 12.6  & 1.23  & 1.46  & 2.3   & 6.1 \\ \hline
Ariadne’s Thread & 3.33  & 2.69  & 4.04  & 4.27  & 4.38  & 3.24  & 2.48  & 2.58  & {NaN}   & {NaN}   & 1.52  & 1.87  & 1.55  & 1.65  & 1.47  & 1.27  & 1.33  & 2.69  & {NaN}   & 2.52 \\ 
\textcolor{black}{TeViA} & \textbf{2.53}  & 2.74  & 3.74  & 3.71  & 4.48  & \textbf{2.22}  & \textbf{2.37}  & \textbf{2.17}  & 4.55  & 2.69  & 1.02  & 1.89  & 1.73  & 1.51  & 1.1   & 1     & 1.15  & 1.35  & 2.3   & 2.35 \\ 
\textcolor{black}{H-SAM} & 5.13  & 4.95  & 6.5   & 6.3   & 8.14  & 5.82  & 5.23  & 5.42  & 4.51  & NaN   & 2.34  & 2.38  & 2.26  & 3.17  & 1.92  & 1.75  & 2.07  & NaN   & NaN   & 3.53 \\ \hline
ATM-Net (Ours)      & 2.79 & \textbf{2.45} & 3.87  & \textbf{3.27} & \textbf{4.01} & 3.53 & 2.61 & 2.49 & \textbf{4.2} & 2.25 & \textbf{1.01} & \textbf{1.16} & \textbf{0.96} & \textbf{1.06} & \textbf{0.85} & \textbf{0.77} & \textbf{0.82} & \textbf{0.68} & \textbf{2.08} & \textbf{2.15} \\
% Ours & 87.18 & 85.08 & 81.64 & 82.1 & 80.29 & 76.26 & 79.25 & 81.04 & 67.03 & 80.18 & 85.01 & 84.45 & 85.25 & 89.13 & 87.96 & 88.11 & 86.14 & 85.55 & 61.02 & 81.72 \\
\hline
\end{tabular}
}
\end{minipage}%
}
\end{minipage}
\end{table*}

%%%%%%%%% FINE-GRAINED RESULTS ON MRSPINESEG 

%%%%%%%%% FINE-GRAINED RESULTS ON SPIDER

\begin{table*}[p]
\centering
\begin{minipage}[b]{0.20\textwidth}
\centering
\rotatebox{90}{%
%%%%% 第一张
\begin{minipage}{1.05\textheight}
\centering
\caption{{\small{DSC (\%) comparisons for all substructures in SPIDER. The best results are highlighted in \textbf{bold}.}}}
\vspace{-0.13cm}
\label{tab:sub-sp-dsc}
\resizebox{1.02\textheight}{!}{%
\renewcommand{\arraystretch}{0.82} 
\begin{tabular}{c|cccccccccccccccccccc}
\hline Method & Spinal Canal & L5 & L4 & L3 & L2 & L1 & T12 & T11 & T10 & T9 & L5/S & L4/L5 & L3/L4 & L2/L3 & L1/L2 & T12/L1 & T11/T12 & T10/T11 & T9/T10 & Avg. \\
\hline U-Net & 75.62 & 74.80  & 61.40  & 56.21 & 51.39 & 58.06 & 60.36 & 51.36 & 42.46 & 0     & 69.18 & 66.18 & 62.52 & 57.21 & 57.56 & 62.37 & 57.22 & 31.74 & 2.18  & 52.52 \\
UNETR & 65.38 & 79.42 & 72.44 & 76.36 & 73.56 & 73.86 & 69.13 & 55.78 & 20.77 & 0     & 71.39 & 75.19 & 81.78 & 79.42 & 78.51 & 75.36 & 62.78 & 31.96 & 0     & 60.16 \\
SegResNet & 75.42 & 81.44 & 76.22 & 76.10  & 70.74 & 74.25 & 73.39 & 64.35 & 0     & 0     & 73.73 & 73.72 & 80.63 & 79.27 & 78.47 & 76.00    & 66.50  & 49.86 & 0     & 61.58 \\
Attention U-Net & 75.26 & 81.47 & 74.26 & 71.27 & 66.98 & 64.47 & 72.74 & 66.58 & 60.40  & 0     & 75.15 & 74.86 & 77.50  & 73.85 & 73.02 & 74.76 & 66.34 & 42.82 & 0     & 62.72 \\
Swin UNETR & 79.02 & 84.56 & 79.82 & 81.08 & \textbf{80.92} & 82.26 & 81.08 & 73.27 & 0     & 0     & 77.28 & 77.76 & 83.09 & 82.87 & 84.30  & \textbf{84} & 76.32 & 59.12 & 0     & 66.67 \\
nnU-NetV2 & 76.55 & 80.96 & 69.23 & 72.16 & 69.31 & 69.55 & 76.33 & 66.99 & 66.10  & 0     & 75.59 & 77.47 & \textbf{84.27} & 82.81 & 83.76 & 83.76 & 76.11 & \textbf{76.55} & 72.72 & 71.59 \\ \hline
 Modified BiSeNet & 77.31 & 80.15 & 74.80 & 77.32 & 72.40  & 75.14 & 72.31 & 65.65 & 55.31 & 0     & 71.94 & 73.48 & 79.28 & 77.44 & 77.18 & 77.76 & 68.10  & 51.20  & 0     & 64.56 \\
U-BiSeNet & 76.94 & 79.87 & 74.35 & 76.35 & 71.66 & 74.07 & 72.28 & 65.66 & 60.64 & 0     & 71.66 & 74.13 & 78.72 & 77.59 & 76.42 & 76.41 & 69.28 & 52.69 & 0     & 64.67 \\
SpineParseNet & 79.20 & \textbf{87.54} & 80.20  & 81.94 & 80.08 & 79.69 & 79.59 & 70.22 & 50.05 & 37.96 & 77.51 & 79.13 & 84.20  & \textbf{84.71} & \textbf{84.81} & 81.96 & 67.63 & 58.32 & 5.87  & 71.08 \\ \hline
Ariadne’s Thread & \textbf{79.23} & 85.36 & \textbf{81.31} & \textbf{83.21} & 80.62 & \textbf{82.66} & 0     & 75.27 & 0     & 0     & 77.08 & 77.87 & 83.06 & 82.09 & 82.44 & 81.97 & 75.63 & 65.77 & 0     & 62.82 \\ 
\textcolor{black}{TeViA} & 78.99 & 83.37 & 79.13 & 80.32 & 79.41 & 80.82 & 79.17 & 75.52 & 74.24 & 73.74 & 73.91 & 76.29 & 81.29 & 81.16 & 81.22 & 80.1  & 77.1  & 67.97 & 73.36 & 77.74 \\ 
\textcolor{black}{H-SAM} & 74.25 & 81.52 & 73.43 & 75.19 & 73.6  & 73.11 & 79.37 & 75.15 & 0     & 0     & \textbf{78.48} & \textbf{83.19} & 83.79 & 83.17 & 83.3  & 83.35 & \textbf{80.00}    & 0     & 0     & 64.01 \\ \hline
ATM-Net (Ours)      & 76.78 & 84.94 & 80.67 & 82.51 & 80.53 & 82.48 & \textbf{81.34} & \textbf{79.15} & \textbf{75.43} & \textbf{76.12} & 76.09 & 77.89 & 83.71 & 82.09 & 82.74 & 82.36 & 78.29 & 67.95 & \textbf{77.37} & \textbf{79.39} \\
% Ours & 87.18 & 85.08 & 81.64 & 82.1 & 80.29 & 76.26 & 79.25 & 81.04 & 67.03 & 80.18 & 85.01 & 84.45 & 85.25 & 89.13 & 87.96 & 88.11 & 86.14 & 85.55 & 61.02 & 81.72 \\
\hline
\end{tabular}
}
\end{minipage}%
}
\end{minipage}%
\hfill
%%%%% 第二张
\begin{minipage}[b]{0.20\textwidth}
\centering
\rotatebox{90}{%
\begin{minipage}{1.05\textheight}
\centering
\caption{{\small{Jaccard (\%) comparisons for all substructures in SPIDER. The best results are highlighted in \textbf{bold}.}}}
\vspace{-0.13cm}
\label{tab:sub-sp-jcc}
\resizebox{1.02\textheight}{!}{%
\renewcommand{\arraystretch}{0.82} 
\begin{tabular}{c|cccccccccccccccccccc}
\hline Method & Spinal Canal & L5 & L4 & L3 & L2 & L1 & T12 & T11 & T10 & T9 & L5/S & L4/L5 & L3/L4 & L2/L3 & L1/L2 & T12/L1 & T11/T12 & T10/T11 & T9/T10 & Avg. \\
\hline U-Net & 65.39 & 63.85 & 49.16 & 43.38 & 39.25 & 45.95 & 48.80  & 39.78 & 30.79 & 0     & 57.39 & 54.05 & 50.05 & 45.39 & 46.87 & 51.81 & 45.33 & 23.97 & 1.14  & 42.23 \\
UNETR & 54.26 & 70.39 & 63.47 & 67.07 & 64.19 & 64.23 & 60.68 & 47.01 & 15.69 & 0     & 60.56 & 64.36 & 72.34 & 70.14 & 69.52 & 65.97 & 52.65 & 26.41 & 0     & 52.05 \\
SegResNet & 64.95 & 73.05 & 67.49 & 67.24 & 61.82 & 65.25 & 65.15 & 54.97 & 0     & 0     & 63.48 & 63.95 & 72.61 & 71.02 & 70.3  & 67.71 & 56.87 & 40.96 & 0     & 54.04 \\
Attention U-Net & 66.15 & 73.23 & 65.79 & 62.28 & 58.08 & 56.42 & 63.99 & 56.98 & 49.93 & 0     & 65.11 & 65.00    & 69.61 & 66.10  & 65.47 & 65.76 & 56.24 & 35.27 & 0     & 54.81 \\
Swin UNETR & 69.88 & 77.16 & 71.65 & 73.05 & 72.30  & 73.88 & 73.52 & 63.67 & 0     & 0     & 67    & 68.03 & 74.96 & 74.86 & 76.26 & 75.27 & 65.93 & 49.57 & 0     & 59.31 \\
nnU-NetV2 & 68.03 & 72.83 & 62.22 & 64.54 & 62.04 & 62.54 & 69.04 & 59.02 & 56.40  & 0     & 65.35 & 67.69 & 76.43 & 75.48 & 76.25 & \textbf{75.62} & 66.37 & \textbf{65.98} & 60.41 & 63.49 \\ \hline
Modified BiSeNet & 68.49 & 70.37 & 65.52 & 67.76 & 63.08 & 65.65 & 63.54 & 55.97 & 45.04 & 0     & 60.93 & 62.62 & 70.04 & 68.42 & 68.6  & 68.63 & 57.66 & 42.22 & 0     & 56.03 \\
U-BiSeNet & 68.15 & 70.36 & 65.15 & 66.99 & 62.40  & 64.48 & 63.55 & 56.16 & 49.86 & 0     & 60.67 & 63.27 & 69.63 & 68.85 & 67.82 & 67.49 & 59.48 & 43.48 & 0     & 56.20 \\
SpineParseNet & \textbf{71.34} & \textbf{80.74} & 72.94 & 74.54 & 72.72 & 72.35 & 72.28 & 61.72 & 40.77 & 26.20  & 68.00 & 69.88 & \textbf{76.75} & \textbf{77.6} & \textbf{77.43} & 73.94 & 57.89 & 49.39 & 3.55  & 63.16 \\ \hline
Ariadne’s Thread & 70.78 & 78.34 & \textbf{73.75} & \textbf{75.41} & \textbf{72.8}  & \textbf{74.92} & 0     & 67.11 & 0     & 0     & 67.9  & 68.35 & 75.64 & 74.88 & 75.31 & 74.30  & 66.06 & 56.43 & 0     & 56.42 \\ 
\textcolor{black}{TeViA} & 70.58 & 75.87 & 71    & 72.19 & 71.35 & 72.74 & 72.39 & 66.62 & 64.39 & 61.46 & 63.88 & 66.27 & 72.98 & 73.39 & 73.7  & 71.77 & 66.99 & 57.45 & 61.18 & 68.75 \\ 
\textcolor{black}{H-SAM} & 65.31 & 72    & 64.78 & 66.24 & 64.84 & 64.81 & 71.43 & 66.34 & 0     & 0     & \textbf{68.95} & \textbf{74.51} & 75.54 & 74.7  & 74.76 & 74.45 & \textbf{69.84} & 0     & 0     & 57.36 \\ \hline
ATM-Net (Ours)      & 67.69 & 77.35 & 72.29 & 74.37 & 72.27 & 74.13 & \textbf{73.79} & \textbf{69.8} & \textbf{65.31} & \textbf{68.28} & 66.13 & 68.35 & 75.9  & 74.38 & 75.21 & 73.97 & 67.66 & 58.46 & \textbf{65.37} & \textbf{70.56} \\
% Ours & 87.18 & 85.08 & 81.64 & 82.1 & 80.29 & 76.26 & 79.25 & 81.04 & 67.03 & 80.18 & 85.01 & 84.45 & 85.25 & 89.13 & 87.96 & 88.11 & 86.14 & 85.55 & 61.02 & 81.72 \\
\hline
\end{tabular}
}
\end{minipage}%
}
\end{minipage}%
\hfill
%%%%% 第三张
\begin{minipage}[b]{0.20\textwidth}
\centering
\rotatebox{90}{%
\begin{minipage}{1.05\textheight}
\centering
\caption{{\small{HD95 (mm) for comparisons all substructures in SPIDER. The best results are highlighted in \textbf{bold}.}}}
\vspace{-0.13cm}
\label{tab:sub-sp-hd95}
\resizebox{1.02\textheight}{!}{%
\renewcommand{\arraystretch}{0.82} 
\begin{tabular}{c|cccccccccccccccccccc}
\hline Method & Spinal Canal & L5 & L4 & L3 & L2 & L1 & T12 & T11 & T10 & T9 & L5/S & L4/L5 & L3/L4 & L2/L3 & L1/L2 & T12/L1 & T11/T12 & T10/T11 & T9/T10 & Avg. \\
\hline 
U-Net & 29.55 & 29.09 & 38.63 & 42.76 & 43.99 & 45.48 & 50.44 & 47.02 & 33.14 & 273.74 & 17.37 & 30.89 & 36.24 & 39.20  & 40.07 & 32.86 & 34.47 & 46.91 & 37.00    & 49.94 \\
UNETR & 38.28 & 15.36 & 15.76 & 12.57 & 15.32 & 17.36 & 19.49 & 22.01 & 30.82 & {NaN} & 7.82  & 7.29  & 9.77  & 8.29  & 9.76  & 9.64  & 11.82 & 12.01 & {NaN} & 15.49 \\
SegResNet & 25.60  & 13.72 & 15.27 & 16.89 & 19.19 & 17.23 & 20.98 & 19.45 & {NaN} & {NaN} & 7.13  & 6.62  & 7.22  & 9.83  & 10.32 & 8.2   & 8.79  & 11.54 & {NaN} & 13.62 \\
Attention U-Net & 22.94 & 24.96 & 18.01 & 18.71 & 20.59 & 19.87 & 20.38 & 19.81 & 17.78 & {NaN} & 7.11  & 9.48  & 9.72  & 12.37 & 10.06 & 9.73  & 11.04 & 13.87 & {NaN} & 15.67 \\
Swin UNETR & 21.91 & 9.95  & 10.03 & 11.00    & 12.10  & 10.75 & 16.55 & 20.67 & {NaN} & {NaN} & 7.58  & \textbf{5.56} & 5.79  & 5.06  & 5.84  & 5.47  & 5.91  & 10.51 & {NaN} & 10.29 \\
nnU-NetV2 & 21.07 & 13.95 & 14.17 & 12.45 & 14.60  & 14.85 & 14.45 & 17.03 & 16.95 & 86.24 & 6.77  & 7.25  & \textbf{4.97} & 4.57  & 4.04  & \textbf{3.97} & 4.85  & \textbf{3.73} & 5.36  & 14.28 \\ \hline
Modified BiSeNet & 15.69 & 25.28 & 16.12 & 24.56 & 18.30  & 11.04 & 12.21 & 12.02 & 14.75 & {NaN} & 6.06  & 9.38  & 8.43  & 7.55  & 6.81  & 6.06  & 5.21  & 8.59  & {NaN} & 12.24 \\
U-BiSeNet & \textbf{15.57} & 14.21 & 13.53 & 12.77 & 16.7  & 12.09 & 11.62 & \textbf{11.33} & 15.10  & {NaN} & 6.58  & 6.24  & 5.41  & 6.57  & 7.74  & 5.68  & \textbf{4.73} & 7.29  & {NaN} & 10.19 \\
SpineParseNet & 23.83 & \textbf{8.02} & \textbf{9.05} & \textbf{9.22} & \textbf{10.13} & 11.01 & 15.01 & 16.90  & 30.85 & 49.83 & \textbf{5.14} & 5.81  & 5.87  & \textbf{4.29} & \textbf{4.01} & 5.55  & 10.24 & 14.73 & 27.39 & 14.05 \\ \hline
Ariadne’s Thread & 19.61 & 10.77 & 11.87 & 10.68 & 12.29 & 11.12 & {NaN}   & 16.68 & {NaN}   & {NaN}   & 7.16  & 7.91  & 7.29  & 9.74  & 6.54  & 7.50   & 7.23  & 7.3   & {NaN}   & 10.25 \\ 
\textcolor{black}{TeViA} & 31.91 & 13.79 & 11.39 & 10.97 & 10.31 & \textbf{9.97}  & 11.64 & 12.32 & 14.63 & 6.03  & 7.53  & 6.83  & 8.55  & 8.49  & 7.23  & 7.19  & 8.28  & 7.85  & 8.07  & 10.68 \\ 
\textcolor{black}{H-SAM} & 32.09 & 15.26 & 14.88 & 15.2  & 14.76 & 15.32 & 16.11 & 18.15 & NaN   & NaN   & 8.49  & 7.98  & 7.94  & 9.85  & 8.25  & 7.43  & 7.96  & NaN   & NaN   & 10.43 \\ \hline
ATM-Net (Ours)      & 30.95 & 12.15 & 11.71 & 10.42 & 11.44 & 12.53 & \textbf{11.50} & 15.63 & \textbf{13.70} & \textbf{5.16} & 6.7   & 5.96  & 5.85  & 7.6   & 5.75  & 5.42  & 4.74  & 7.55  & \textbf{3.44} & \textbf{9.91} \\
% Ours & 87.18 & 85.08 & 81.64 & 82.1 & 80.29 & 76.26 & 79.25 & 81.04 & 67.03 & 80.18 & 85.01 & 84.45 & 85.25 & 89.13 & 87.96 & 88.11 & 86.14 & 85.55 & 61.02 & 81.72 \\
\hline
\end{tabular}
}
\end{minipage}%
}
\end{minipage}%
\hfill 
%%%%% 第四张
\begin{minipage}[b]{0.20\textwidth}
\centering
\rotatebox{90}{%
\begin{minipage}{1.05\textheight}
\caption{{\small{ASD (mm) comparisons for all substructures in SPIDER. The best results are highlighted in \textbf{bold}.}}}
\vspace{-0.13cm}
\centering
\label{tab:sub-sp-asd}
\resizebox{1.02\textheight}{!}{%
\renewcommand{\arraystretch}{0.82} 
\begin{tabular}{c|cccccccccccccccccccc}
\hline Method & Spinal Canal & L5 & L4 & L3 & L2 & L1 & T12 & T11 & T10 & T9 & L5/S & L4/L5 & L3/L4 & L2/L3 & L1/L2 & T12/L1 & T11/T12 & T10/T11 & T9/T10 & Avg. \\
\hline 
U-Net & 7.02  & 7.61  & 11.31 & 14.00    & 14.98 & 14.94 & 14.45 & 15.46 & 12.10  & 236.24 & 5.88  & 9.31  & 12.82 & 14.73 & 14.6  & 12.09 & 10.67 & 34.09 & 13.43 & 25.04 \\
UNETR & 5.64  & 3.13  & 3.35  & 2.86  & 3.55  & 4.06  & 4.90   & 7.21  & 3.65  & {NaN} & 2.50   & 2.44  & 2.83  & 2.54  & 3.52  & 3.82  & 5.93  & 4.73  & {NaN} & 3.92 \\
SegResNet & 6.57  & 2.86  & 3.94  & 4.55  & 4.89  & 4.49  & 4.89  & 5.08  & {NaN} & {NaN} & 2.87  & 3.03  & 3.17  & 3.85  & 4.47  & 4.02  & 3.85  & 4.03  & {NaN} & 4.16 \\
Attention U-Net & 2.75  & 6.22  & 4.73  & 4.98  & 5.27  & 4.80   & 4.69  & 4.56  & 2.96  & {NaN} & 2.53  & 3.75  & 3.79  & 4.74  & 4.42  & 4.17  & 4.68  & 6.48  & {NaN} & 4.44 \\
Swin UNETR & 5.41  & 2.42  & 2.95  & 3.28  & 3.11  & 2.96  & 4.91  & 6.18  & {NaN} & {NaN} & 2.64  & 2.65  & 2.75  & 2.45  & 2.44  & 1.81  & 2.19  & 3.12  & {NaN} & 3.21 \\
nnU-NetV2 & 3.47  & 2.96  & 2.45  & \textbf{2.32} & 3.15  & 3.29  & 2.98  & 3.69  & \textbf{2.54} & 43.76 & 2.13  & 2.35  & \textbf{1.74} & \textbf{1.75} & \textbf{1.6} & \textbf{1.46} & 2.19  & \textbf{1.05} & 0.98  & 4.52 \\ \hline
Modified BiSeNet & 2.79  & 6.01  & 3.66  & 4.86  & 4.94  & 2.71  & 3.68  & 3.67  & 2.96  & {NaN} & 2.24  & 2.95  & 3.07  & 2.8   & 3.37  & 2.93  & 2.43  & 2.69  & {NaN} & 3.40 \\
U-BiSeNet & 2.93  & 3.42  & 3.51  & 3.28  & 4.25  & 2.97  & 3.45  & \textbf{3.26} & 3.61  & {NaN} & 2.47  & 2.72  & 2.65  & 2.84  & 3.63  & 2.77  & 2.02  & 2.64  & {NaN} & 3.08 \\
SpineParseNet & \textbf{2.64} & \textbf{1.72} & \textbf{1.78} & 2.95  & \textbf{2.29} & \textbf{1.9} & \textbf{2.84} & 4.65  & 4.54  & 1.33  & \textbf{1.94} & \textbf{2.23} & 3.45  & 2.25  & 2.09  & 2.34  & 4.13  & 8.66  & 6.14  & 3.15 \\ \hline
Ariadne’s Thread & 3.85  & 2.82  & 3.26  & 3.09  & 3.14  & 3.12  & {NaN}   & 5.11  & {NaN}   & {NaN}   & 2.63  & 3.10  & 3.22  & 3.87  & 3.10  & 2.99  & 3.23  & 1.23  & {NaN}   & 3.18  \\ 
\textcolor{black}{TeViA} & 7.76  & 3.01  & 3.16  & 3.12  & 3.2   & 3.23  & 3.44  & 4.65  & 3.04  & 2.72  & 2.67  & 2.62  & 2.52  & 2.33  & 2.22  & 2.17  & 3.07  & 2.51  & 2.45  & 3.15 \\ 
\textcolor{black}{H-SAM} & 5.37  & 4.16  & 3.75  & 4.3   & 4.07  & 3.92  & 3.93  & 5.31  & NaN   & NaN   & 2.94  & 3.23  & 2.84  & 2.78  & 2.86  & 2.62  & 2.81  & NaN   & NaN   & 2.84 \\ \hline
ATM-Net (Ours)      & 6.91  & 2.87  & 2.89  & 2.83  & 2.86  & 3.06  & 2.95  & 4.77  & 3.15  & \textbf{0.93} & 2.51  & 2.50   & 2.57  & 3.05  & 2.51  & 2.47  & \textbf{1.72} & 1.18  & \textbf{0.96} & \textbf{2.77} \\
% Ours & 87.18 & 85.08 & 81.64 & 82.1 & 80.29 & 76.26 & 79.25 & 81.04 & 67.03 & 80.18 & 85.01 & 84.45 & 85.25 & 89.13 & 87.96 & 88.11 & 86.14 & 85.55 & 61.02 & 81.72 \\
\hline
\end{tabular}
}
\end{minipage}%
}
\end{minipage}
\end{table*}

\subsubsection{Quantitative results (Overall)} 

{\color{black}In the left side of Table~\ref{tab:comprehensive-compare}, we present the comparative results on the MRSpineSeg dataset and SPIDER dataset, including general-purpose MIS models (\scalebox{1}{$\bullet$}), lumbar spine-specific segmentation models (\scalebox{0.8}{$\blacktriangle$}), Visual-Language Model (VLM) / Visual Foundation Model (VFM) (\scalebox{1}{$\star$}), and the proposed ATM-Net.
From this table, we have the following observations.
}

(1) \textit{\textbf{ATM-Net achieves SOTA performance against all MIS benchmarks on both datasets.}}
% \noindent \textbf{Considering 
Compared to the classic MIS benchmark U-Net, ATM-Net achieves immense improvements across all metrics. For instance, on MRSpineSeg, ATM-Net enhances the DSC↑ and Jaccard↑ by 23.13\% and 24.49\%, respectively, and decreases HD95↓ and ASD↓ by 23.05 and 6.2 pixels, respectively.
Compared to the powerful nnUNetV2 benchmark, our method shows marked and consistent enhancements on SPIDER, increasing DSC by 7.80\% and decreasing HD95 by 4.37 pixels. These results demonstrate the model's excellence in both regional overlap and boundary accuracy. 
% Similarly, on the SPIDER, our method improves DSC by 7.80\% and reduces HD95 by 4.37, reaffirming ATM-Net's exceptional capability in the challenging task of lumbar spine segmentation.

% \thispagestyle{empty}

(2) \textit{\textbf{ATM-Net demonstrates impressive superiority over lumbar spine-specific segmentation models.}}
On both datasets, ATM-Net shows consistent improvements across all the metrics even compared with the powerful SpineParseNet. For example, in MRSpineSeg, ATM-Net surpasses SpineParseNet by 2.88\% and 0.6\% in DSC and Jaccard, respectively. 
Additionally, ATM-Net significantly reduces the HD95 and ASD metrics by 7.94 and 3.95 pixels, indicating improved accuracy in edge detection.
Similarly, on the SPIDER dataset, ATM-Net shows an enhancement in DSC and Jaccard scores by 8.31\% and 7.40\%, while also decreasing the HD95 and ASD values by 4.14 and 0.38 pixels, respectively. 
These results confirm its robustness in capturing nuanced lumbar spine anatomy.
% These results underscore ATM-Net's ability to provide more accurate and detailed segmentation, especially beneficial for the nuanced anatomy of the lumbar spine.

{\color{black}
(3) \textit{\textbf{ATM-Net outperforms the representative VLM / VFM.}}
A-Thread, TeViA, and H-SAM, as the representative VLM / VFM solution, falls short of ATM-Net in this specialized lumbar spine segmentation task.
For example, on SPIDER, ATM-Net outperforms A-Thread by 16.57\% in DSC, 14.14\% in Jaccard, and reduces HD95 and ASD by 0.34 and 0.41 pixels, respectively.
These results highlight the limitations of general VLM / VFM approaches in capturing the fine-grained details required for lumbar spine segmentation and validate the effectiveness of our tailored multimodal design.
}

% (4) \textit{Compared to the Swin UNETR model with the same image encoder, ATM-Net consistently achieves remarkable improvements across various metrics, highlighting the significant advantages of incorporating clinical text information and the multimodal fusion mechanism within ATM-Net.}
(4) \textit{\textbf{ATM-Net outperforms Swin UNETR, demonstrating the advantage of incorporating text insights.}}
% Compared to Swin UNETR with the same image encoder, ATM-Net consistently shows significant improvements, demonstrating the benefits of integrating textual insights.}}
Sharing the same visual encoder as Swin UNETR, ATM-Net’s additional text insights and fusion mechanism lead to remarkable boosts.
On MRSpineSeg, it reduces HD95 by 7.72 pixels and ASD by 4.89 pixels, while on SPIDER, DSC and Jaccard increase by 12.72\% and 11.25\%, respectively. 
% This underscores the value of incorporating clinical text insights to enhance fine-grained segmentation of lumbar spine structures.

% As mentioned in \Cref{subsec:encoder}, ATM-Net adopts the Swin UNETR as its visual encoder. In addition to the visual branch, ATM-Net further incorporates crucial text information so that the overall performance witnesses remarkable boosts.
% For example, on MRSpineSeg, ATM-Net excels in terms of HD95 (9.6) and ASD (2.15),  while Swin UNETR obtained a HD95 of 17.32 and an ASD of 7.04.
% On SPIDER, ATM-Net achieves the DSC of 79.39\% and the Jaccard of 70.56\%, significantly surpassing the ones of Swin UNETR by 12.72\% and 11.25\%, respectively.
% on MRSpineSeg, ATM-Net achieves the DSC of 81.72\% and the Jaccard of 72.25\%, significantly surpassing the ones of Swin UNETR by 17.14\% and 15.47\%, respectively.

These comparative results demonstrate that our method's integration of clinical text information into the existing image branch significantly enhances the model's performance in fine-grained lumbar spine segmentation tasks. This improvement is evident in both overall segmentation quality and boundary delineation accuracy.

% further introduces a textual branch to incorporate crucial clinical text priors at different levels. This multimodal information is then adaptively fused, which significantly enhances the segmentation performance.

\begin{figure*}[h]
    \centering
    \includegraphics[width=1.05\linewidth]{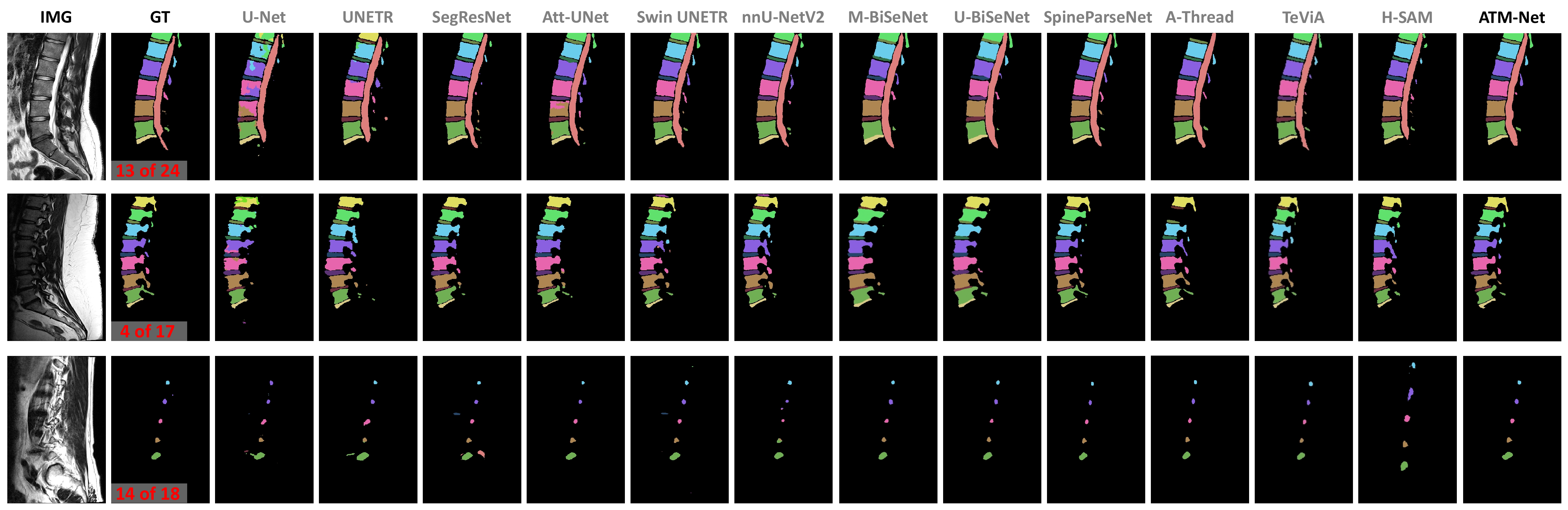}
    \caption{\textcolor{black}{Qualitative comparison of ATM-Net with other methods across different sagittal planes. Best viewed in color.}}
    \label{fig:qualitative-all}
\end{figure*}

\subsubsection{Quantitative results (considering all substructures)} 
We also present the segmentation results of ATM-Net and other comparative methods across two datasets for each fine-grained tissue structure category, evaluated using the same four metrics.
The detailed results are shown in \Cref{tab:sub-mr-dsc} to \Cref{tab:sub-sp-asd}.
% Specifically, the DSC comparing results on MRSpineSeg is listed in Table~\ref{tab:sub-mr-dsc}.
From these tables, we observe that:

1) \textit{\textbf{ATM-Net achieves superior or comparative results across all the fine-grained substructures.}}
For example, among all the 19 substructures on MRSpineSeg, ATM-Net achieved the highest DSC in 7 substructures, while demonstrating competitive results in the remaining 12 categories. 
Meanwhile, ATM-Net achieved the best HD95 in 12 substructures and demonstrated competitive results in the other 7 categories.
% Similarly, among all the 19 substructures on SPIDER, ATM-Net achieved the best HD95 in 18 substructures, while demonstrating competitive results in the remaining category (\textit{L4/L5}). 
These outstanding results have also led ATM-Net to achieve the best overall performance.

2) \textit{\textbf{ATM-Net demonstrates exceptional performance on the challenging categories.}}
Due to factors such as class imbalance (illustrated in \Cref{fig:class-distribution}) and class similarity, many comparing methods struggle in discriminating some challenging substructures,
% , such as \textit{T9,} and \textit{T9/T10}.
whereas ATM-Net demonstrates stable performance. % in these challenging categories. 
For example,  \textit{T9/T10} is the least frequently occurring ID across all data. In this challenging category from the MRSpineSeg dataset, ATM-Net achieved a DSC of 61.02, significantly surpassing the second-best solution (SpineParseNet) by 5.93\%. Meanwhile, most methods yield a DSC of 0 in this category.
This can be primarily \textbf{\textit{attributed to two key factors}}: First, the text insights introduced by ATM-Net exhibit robustness to class distribution variations. Specifically, its CCAE module strengthens multi-modal consistency for low-frequency classes and maximizes mutual information across different modalities, thereby substantially improving segmentation accuracy for these challenging categories. Second, ATM-Net incorporates Focal loss (\Cref{subsec:HASF}), which also effectively alleviates the class imbalance issue.

\subsubsection{Complexity analysis}
We further conducted complexity analysis and the comparative results are illustrated on the right side of \Cref{tab:comprehensive-compare}, which reports three key metrics: the number of parameters, FLOPs, and inference time.

{\color{black}
ATM-Net achieves an inference speed of 75ms per slice, which is only ~14\% slower than the baseline Swin UNETR (65.87 ms). 
We need to clarify that, the reported 75ms refers to the model forward propagation, including the visual backbone and fusion modules (HASF + CCAE). Text prompt generation and BERT encoding are performed offline as one-time operations, and their cost is negligible when amortized across all slices in a volume.}
Meanwhile, it delivers substantial accuracy gains: on the SPIDER dataset, its Dice coefficient increases from 66.67\% (Swin UNETR) to 79.39\%, corresponding to an absolute improvement of 12.72\%. Compared with the VLM baseline Ariadne’s Thread, ATM-Net also has fewer parameters (64.87 M vs. 115.39 M) and lower FLOPs (623.95 G vs. 634.2 G). These results confirm that ATM-Net effectively balances segmentation performance and inference efficiency, achieving significant accuracy improvements while maintaining reasonable computational complexity.

% As shown in the Table.~\ref{tab:complexity}, our method achieves an inference speed of 75 ms per slice, which is only about a 14\% increase compared to the baseline Swin UNETR (65.87 ms). However, as reported in Table.~\ref{tab:compare-exp}, our method yields a remarkable improvement in segmentation accuracy. For instance, on the SPIDER dataset, the Dice coefficient increases from 66.67\% to 79.39\%. Furthermore, compared with other vlm such as Ariadne’s Thread, our method requires fewer parameters and FLOPs. These results demonstrate that the proposed model effectively balances segmentation performance and inference efficiency, achieving significant accuracy gains while maintaining reasonable computational complexity.

\begin{figure*}[h]
    \centering
    \includegraphics[width=0.93\linewidth]{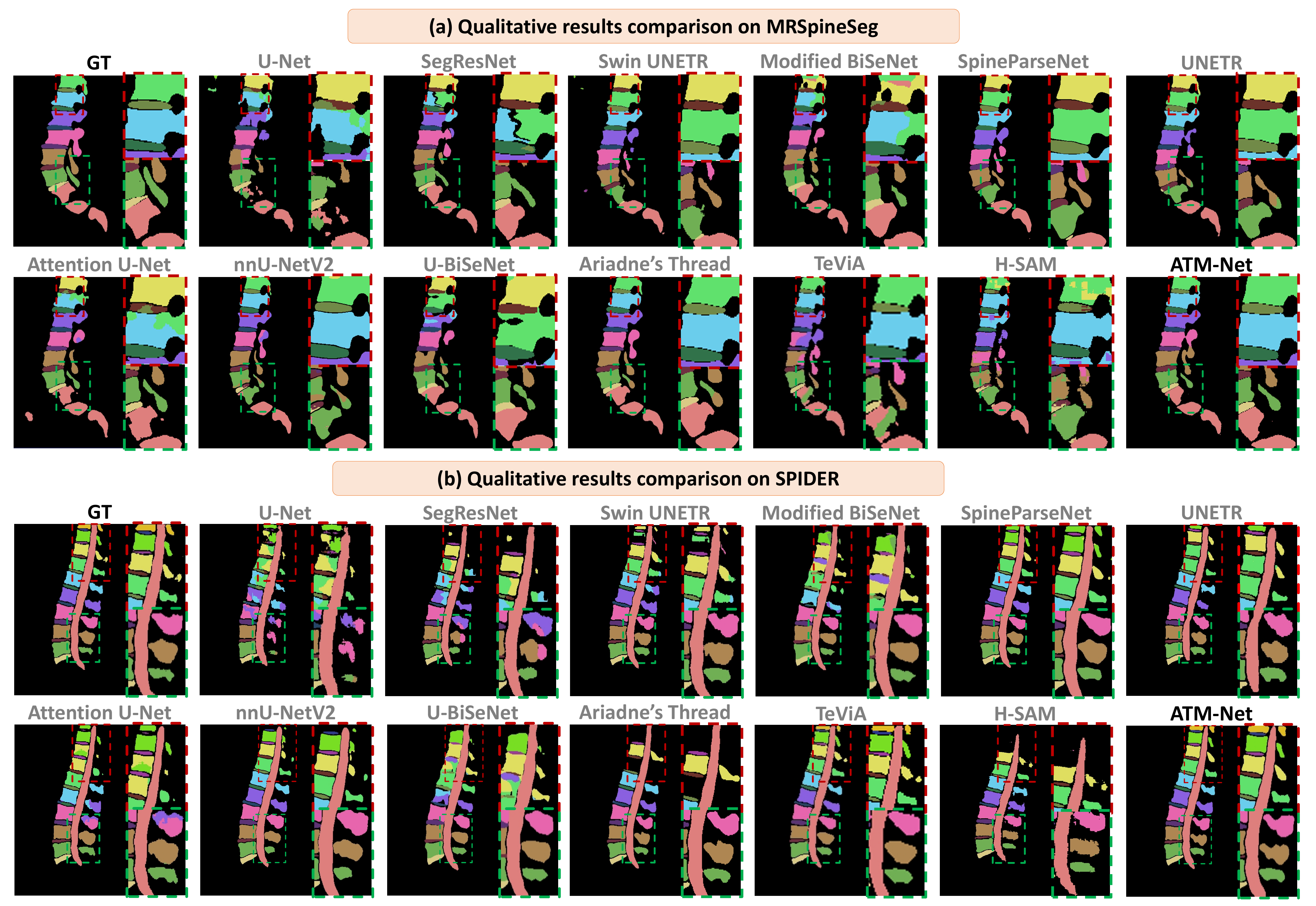}
    \caption{\textcolor{black}{Qualitative comparison of segmentation results between ATM-Net and other methods across two datasets. We also provide zoomed-in views, with dashed boxes: red for class discrimination and green for segmentation details. Best viewed in color.}}
    \label{fig:qualitative-detail}
\end{figure*}

\subsubsection{Qualitative results}
We present qualitative comparisons of segmentation predictions across different sagittal planes in \Cref{fig:qualitative-all}. As shown in the figure, our ATM-Net demonstrates significant advantages over the comparison methods in both distinguishing similar categories and capturing segmentation details.
In \Cref{fig:qualitative-detail}, we further provide a comparison of segmentation predictions for a specific slice in both datasets, along with zoom-in views. 
% The red dashed boxes focus on category discrimination, while the green dashed boxes highlight segmentation details.
From the red dashed boxes, it can be observed that classic MIS methods such as UNet, Swin UNETR, spine-specific solutions such as U-BiseNet and SpineParseNet, and VLM baseline Ariadne's Thread, struggle to differentiate between close and challenging categories such as \textit{T12, T12/L1} and \textit{L1}. However, ATM-Net exhibits excellent performance in distinguishing these categories. A similar situation is evident in the segmentation details outlined by the green dashed boxes, where nearly all comparison methods struggle to provide precise predictions for some small and challenging structures, while ATM-Net yields satisfactory predictions.
These factors demonstrate the efficacy of encoded anatomical text insights in ATM-Net.

\begin{figure}
    \centering
    \includegraphics[width=0.68\linewidth,height=7.2cm]{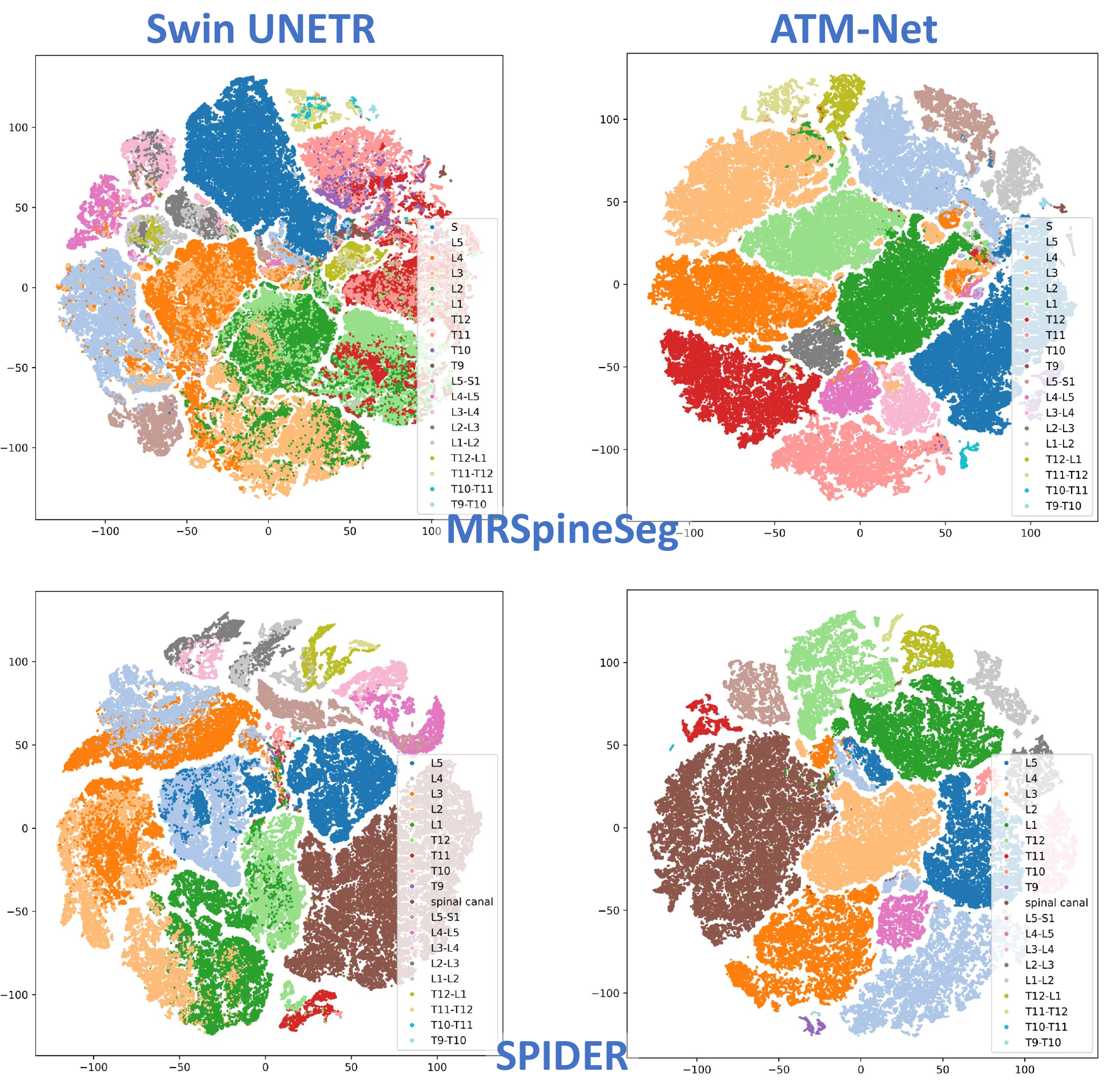}
    \caption{The t-SNE visualization of embedding space on both datasets for Swin UNETR and our ATM-Net. Best viewed in color.}
    \label{fig:tsne}
\end{figure}

To further explore the efficacy of ATM-Net, we exhibit the t-SNE visualization of embedding space for Swin-UNETR and ATM-Net (\Cref{fig:tsne}).
We observe that ATM-Net notably shows better feature clustering and anatomical structure discrimination in both datasets. 
For example, in MRSpineSeg, the features of \textit{L1, L2,} and \textit{L3} are mixed with each other due to factors such as similar appearance and imbalanced category distribution.
After integrating powerful anatomical text insights, our ATM-Net is capable of accurately distinguishing these challenging categories and demonstrates significantly clearer classification boundaries.

% \begin{table*}[]
%    \centering
% \caption{Ablation study results regarding HASF and CCAE module. The best results are highlighted in bold.}
% \resizebox{0.98\linewidth}{!}{
% \begin{tabular}{cc|cccc|cccc}
% \hline \multicolumn{2}{c}{ Module } & \multicolumn{4}{|c|}{ MRSpineSeg } & \multicolumn{4}{c}{ SPIDER } \\
% \hline HASF & CCAE & DSC (\%) $\uparrow$ & Jaccard (\%) $\uparrow$ & HD95 (pixel) $\downarrow$ & ASD (pixel) $\downarrow$ & DSC (\%) $\uparrow$ & Jaccard (\%) $\uparrow$ & HD95 (pixel) $\downarrow$ & ASD (pixel) $\downarrow$ \\
% \hline & & 64.58 & 56.78 & 17.32 & 7.04 & 66.67 & 59.31 & 10.29 & 3.21 \\
% & $\checkmark$ & 73.08 & 64.37 & 13.43 & 3.52 & 69.23 & 61.01 & 12.97 & 4.02 \\
% $\checkmark$ & & 77.13 & 68.46 & 10.10 & 2.80 & 73.00 & 64.04 & 12.07 & 3.54 \\
% $\checkmark$ & $\checkmark$ & \textbf{81.72} & \textbf{72.25} & \textbf{9.60} & \textbf{2.15} & \textbf{79.39} & \textbf{70.56} & \textbf{9.91} & \textbf{2.77} \\
% \hline
% \end{tabular}
% }
% \label{tab:ablation}
% \end{table*}

\begin{table*}[htbp]
  \centering
  \caption{Ablation study results regarding HASF and CCAE module. The best results are highlighted in bold, \textcolor{black}{and '*' denotes statistical significance ($p < 0.05$).}}
  \label{tab:ablation}
  
  \renewcommand\arraystretch{1.2} % 增加行间距
  \footnotesize % 使用更小字体以节省空间
  \begin{tabularx}{\linewidth}{
    >{\centering\arraybackslash}X  % HASF列（自适应宽度）
    >{\centering\arraybackslash}X|  % CCAE列（自适应宽度）
    >{\centering\arraybackslash}X  % MRSpineSeg DSC
    >{\centering\arraybackslash}X  % MRSpineSeg Jaccard
    >{\centering\arraybackslash}X  % MRSpineSeg HD95
    >{\centering\arraybackslash}X|  % MRSpineSeg ASD
    >{\centering\arraybackslash}X  % SPIDER DSC
    >{\centering\arraybackslash}X  % SPIDER Jaccard
    >{\centering\arraybackslash}X  % SPIDER HD95
    >{\centering\arraybackslash}X   % SPIDER ASD
  }
    \toprule
    
    \multicolumn{2}{c|}{\textbf{Module}} & 
    \multicolumn{4}{c|}{\textbf{MRSpineSeg}} & 
    \multicolumn{4}{c}{\textbf{SPIDER}} \\
    
    \cmidrule{1-10}
    
    \textbf{HASF} & \textbf{CCAE} & 
    \makecell{DSC\\(\%)$\uparrow$} & \makecell{Jacc.\\(\%)$\uparrow$} & \makecell{HD95\\(px)$\downarrow$} & \makecell{ASD\\(px)$\downarrow$} &
    \makecell{DSC\\(\%)$\uparrow$} & \makecell{Jacc.\\(\%)$\uparrow$} & \makecell{HD95\\(px)$\downarrow$} & \makecell{ASD\\(px)$\downarrow$} \\
    
    \midrule
    
     &  & 64.58* & 56.78* & 17.32* & 7.04* & 66.67* & 59.31* & 10.29* & 3.21* \\
     & $\checkmark$ & 73.08* & 64.37* & 13.43* & 3.52* & 69.23* & 61.01* & 12.97* & 4.02* \\
    $\checkmark$ &  & 77.13* & 68.46* & 10.10 & 2.80* & 73.00* & 64.04* & 12.07* & 3.54* \\
   \rowcolor{gray!20} $\checkmark$ & $\checkmark$ & \textbf{81.72} & \textbf{72.25} & \textbf{9.60} & \textbf{2.15} & \textbf{79.39} & \textbf{70.56} & \textbf{9.91} & \textbf{2.77} \\
    
    \bottomrule
  \end{tabularx}
\end{table*}

\begin{table}[htbp]
  \centering
  \caption{\textcolor{black}{Ablation study of different text injection strategies within the HASF module. The best results are highlighted in bold.}}
    \begin{tabular}{l|cccc}
    \toprule
    Method & DSC (\%)$\uparrow$ & Jacc. (\%)$\uparrow$ & HD95 (pixels)$\downarrow$ & ASD (pixels)$\downarrow$ \\
    \hline
    Bottleneck-only injection
 & 76.25 & 67.61 & 9.66 & 2.36 \\
    All-stage injection
 & \textbf{81.72} & \textbf{72.25} & \textbf{9.60} & \textbf{2.15} \\
    \bottomrule
    \end{tabular}%
  \label{tab:injection_ablation}%
\end{table}%

\begin{table}[htbp]
  \centering
  \caption{\textcolor{black}{Ablation study on the hyperparameter $\lambda_2$ in the loss function. The best results are highlighted in bold.}}
    \begin{tabular}{l|cccc}
    \toprule
    $\lambda_2$ & DSC (\%)$\uparrow$ & Jacc. (\%)$\uparrow$ & HD95 (pixels)$\downarrow$ & ASD (pixels)$\downarrow$ \\
    \hline
    0.1 & 76.91 & 68.05 & 10.46 & 2.99 \\
    0.2 & \textbf{81.72} & \textbf{72.25} & \textbf{9.60} & \textbf{2.15} \\
    0.3 & 81.04 & 71.71 & 10.28 & 2.73 \\
    \bottomrule
    \end{tabular}%
  \label{tab:lambda_sensitivity}%
\end{table}%

\subsection{Ablation study}
\label{subsec:ablation}

\subsubsection{Effectiveness of HASF and CCAE}

We investigate the contribution of the key components in ATM-Net, e.g., HASF and CCAE, and demonstrate the ablation results on both datasets in Table~\ref{tab:ablation}.
From this table,  we have the following observations.

\textit{\textbf{(1) HASF significantly boosts the overall segmentation performance of ATM-Net.}} 
The proposed ATM-Net shows substantial boosts on both datasets by integrating holistic anatomical textual information through the HASF module. Compared with the baseline, when incorporating HASF, the overall DSCs rise from 64.58\% to 77.13\% on MRSpineSeg, and from 66.67\% to 73.00\% on SPIDER, respectively.

\textcolor{black}{Furthermore, we evaluate the text injection strategy within HASF (Table~\ref{tab:injection_ablation}). Compared to ``Bottleneck-only injection", our ``All-stage injection" significantly improves performance, reaching a DSC of 81.72\% (+5.47\%) and an ASD of 2.15 pixels. This confirms that injecting textual semantics across multiple decoder stages is essential, as it allows anatomical priors to simultaneously guide high-level semantic localization and low-level boundary delineation.}

\textit{\textbf{(2) CCAE also brings consistent improvement towards ATM-Net.}}
When integrating CCAE, ATM-Net adaptively relieves inter-class similarity issues through channel-wise contrastive learning.
For instance, compared with the baseline, when incorporating CCAE, the overall Jaccards improve by 7.59\% on MRSpineSeg, and 1.70\% on SPIDER, respectively. 

\textcolor{black}{
To investigate the influence of ATM-Net to the hyperparameters in the overall loss function (Eq. 8), we conducted a systematic ablation study on the weighting factor $\lambda_2$, which controls the auxiliary contrastive loss $\mathcal{L}_{ftc}$, while keeping the primary segmentation loss weight $\lambda_1$ fixed at 1. As demonstrated in \Cref{tab:lambda_sensitivity}, the model achieves its optimal segmentation performance when $\lambda_2$ is set to 0.2. When $\lambda_2$ is too small (e.g., 0.1), the penalty is insufficient to effectively guide the cross-modal feature alignment, leading to a performance drop. Conversely, when $\lambda_2$ is too large (e.g., 0.3), the auxiliary loss potentially overshadows the primary segmentation loss, which also results in suboptimal performance. Therefore, we set the hyperparameter $\lambda_2$ to $0.2$.
}

\textit{\textbf{(3) When incorporating both HASF and CCAE, ATM-Net achieves the most optimal results.}} As indicated in the last row of Table~\ref{tab:ablation}, when incorporating both HASF and CCAE, ATM-Net achieves consistent and significant improvement across all the metrics on both datasets.
Both DSC and Jaccard witness a boost of at least 10\% on both datasets.
Considering metrics regarding boundary similarity, ATM-Net gets comparative results with only one of the components (discussed in \Cref{sec:discuss}(1)). However, when integrating both HASF and CCAE, ATM-Net achieves consistent improvements.
The ASD drops by 4.89 and 0.44 on MRSpineSeg and SPIDER, respectively. These factors demonstrate that integrating both holistic and channel-wise anatomical textual information is crucial for achieving robust fine-grained segmentation of lumbar spine MRI.

\begin{figure}
    \centering
    \includegraphics[width=0.93\linewidth]{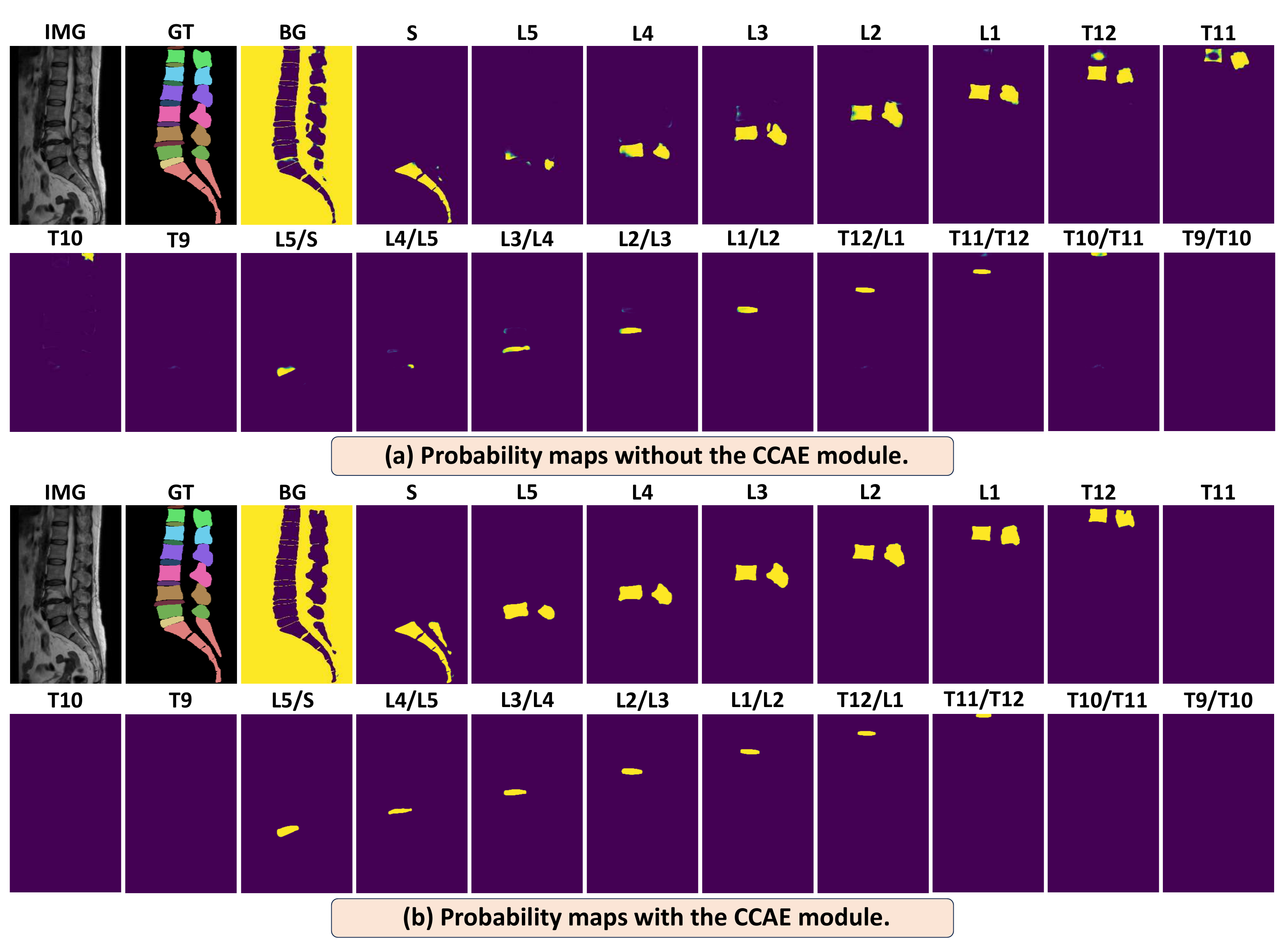}
    \caption{ Visual comparison of the predicted probability maps. (a) Results obtained without incorporating the CCAE module. (b) Results obtained with the proposed CCAE module.}
    \label{fig:probability_map}
\end{figure}

To further validate the effectiveness of the proposed CCAE module, we provide visual comparisons of the predicted probability maps with and without CCAE. As shown in \Cref{fig:probability_map}, the incorporation of CCAE enables the model to better overcome inter-class similarities and more accurately distinguish adjacent vertebrae and intervertebral discs. For instance, without the CCAE module, the model fails to accurately distinguish the vertebrae L5 and L4, whereas after introducing CCAE, these structures can be more clearly and accurately separated; a similar improvement is observed for the intervertebral discs L5/S and L4/L5.
Moreover, the confidence along the target boundaries increases noticeably with CCAE. These results demonstrate that CCAE effectively enhances feature discrimination among neighboring anatomical structures, allowing the network to produce more discriminative and reliable probability maps, thereby leading to more accurate segmentation results.

\textcolor{black}{To validate the architectural choice of the text encoder in our proposed framework, we conducted an ablation study evaluating three different pre-trained language models on the MRSpineSeg dataset: a general-domain encoder (RoBERTa) and two medical-domain encoders (PubMedBERT and Bio-ClinicalBERT). The quantitative results are presented in Table~\ref{tab:text_encoder_ablation}. As observed, Bio-ClinicalBERT demonstrates the most optimal overall segmentation performance, yielding the highest DSC (81.72\%) and Jaccard index (72.25\%), as well as the lowest ASD (2.15 px). Although the general-domain encoder RoBERTa achieves a marginally better HD95 score (9.52 px compared to 9.60 px), it significantly underperforms in region-overlap metrics, with a DSC drop of nearly 4\%. Furthermore, compared to PubMedBERT, Bio-ClinicalBERT exhibits consistent superiority across all evaluation metrics. This superior performance can be primarily attributed to Bio-ClinicalBERT's specialized pre-training on large-scale clinical corpora (e.g., MIMIC-III). This endows the model with a profound capability to capture intricate medical semantics, ensuring a highly accurate semantic alignment with the specific anatomical terminology (e.g., ``lumbar vertebra'', ``sagittal plane'') utilized in our prompts. Consequently, Bio-ClinicalBERT is selected as the default text encoder for ATM-Net.}

\begin{table}[htbp]
  \centering
  \caption{\textcolor{black}{Ablation study of text encoder on the MRSpineSeg Dataset. The best results are highlighted in bold.}}
    \begin{tabular}{l|cccc}
    \toprule
    Text Encoder & DSC (\%)$\uparrow$ & Jacc. (\%)$\uparrow$ & HD95 (pixels)$\downarrow$ & ASD (pixels)$\downarrow$ \\
    \hline
    RoBERT & 77.78 & 69.10 & \textbf{9.52} & 2.18 \\
    PubMedBERT & 81.42 & 72.04 & 10.65 & 2.80 \\
    Bio-ClinicalBERT & \textbf{81.72} & \textbf{72.25} & 9.60 & \textbf{2.15} \\
    \bottomrule
    \end{tabular}%
  \label{tab:text_encoder_ablation}%
\end{table}%

\subsubsection{The granularity of the textual prompt in HASF}
\label{subsub:granularity}

The choice of textual prompt is also an important factor in ATM-Net.
We conducted ablation experiments on the choice of granularity on the textual prompts used in the HASF module.

\begin{figure*}
    \centering
    \includegraphics[width=0.95\linewidth]{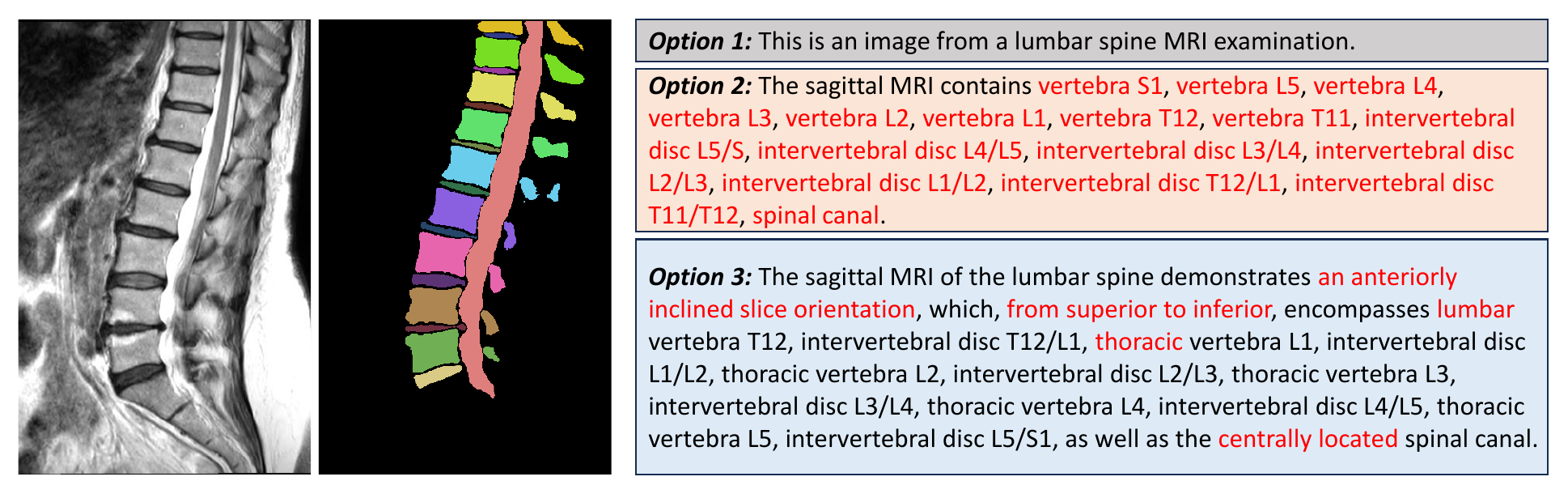}
    \caption{Different prompt selections: from Option 1 to Option 3, the granularity of image descriptions varies from coarse to fine. Best viewed in color.}
    \label{fig:prompt-opt}
\end{figure*}

\Cref{fig:prompt-opt} visually presents examples of different choices of textual prompts. From \textit{Option 1} to \textit{Option 3},  the granularity of the textual information gradually increases. Specifically, \textit{Option 1} only describes the overall type of the image, while \textit{Option 2} lists different structures within the image. \textit{Option 3} further enhances this by adding information about the slice's positional context and the spatial relationships between different structures, with the description being closer to clinical diagnostic reports.

The experimental results regarding prompt choices across both datasets are detailed in Table~\ref{tab:prompt-opt}. 
Through this table, we have the following observations.

\begin{table*}[htbp]
  \centering
  \caption{The ablation study results for different prompt options.}
  \label{tab:prompt-opt}
  
  \small
  \begin{tabularx}{\linewidth}{c|*{4}{>{\centering\arraybackslash}X}|*{4}{>{\centering\arraybackslash}X}}
    \toprule
    
    \multirow{2}{*}{\textbf{Method}} & 
    \multicolumn{4}{c|}{\textbf{MRSpineSeg}} & 
    \multicolumn{4}{c}{\textbf{SPIDER}} \\
    
    \cmidrule{2-9}
    
    & \makecell{DSC\\{\tiny{(\%)$\uparrow$}}} & \makecell{Jacc.\\{\tiny{(\%)$\uparrow$}}} & \makecell{HD95\\{\tiny{(pixel)$\downarrow$}}} & \makecell{ASD\\{\tiny{(pixel)$\downarrow$}}} 
    & \makecell{DSC\\{\tiny{(\%)$\uparrow$}}} & \makecell{Jacc.\\{\tiny{(\%)$\uparrow$}}} & \makecell{HD95\\{\tiny{(pixel)$\downarrow$}}} & \makecell{ASD\\{\tiny{(pixel)$\downarrow$}}} \\
    
    \midrule
    
    w/o text   & 64.58 & 56.78 & 17.32 & 7.04 & 66.67 & 59.31 & 10.29 & 3.21 \\
    Option 1   & 70.28 & 62.45 & 10.22 & 2.68 & 69.54 & 61.93 & 11.06 & 3.07 \\
    Option 2   & 75.80 & 66.84 & 11.28 & 3.32 & 71.64 & 64.04 & 9.61 & 2.78 \\
    Option 3   & 77.13 & 68.46 & 10.10 & 2.80 & 73.00 & 64.04 & 12.07 & 3.54 \\
    
    \bottomrule
  \end{tabularx}
\end{table*}

\textit{\textbf{(1) ATM-Net consistently improves with any text prompt granularity.}}
No matter integrating \textit{Opt.1, Opt.2}, or \textit{Opt.3} with varying levels of granularity, ATM-Net achieves consistent improvements. For example, when integrating one of these options, ASD$\downarrow$ in MRSPineSeg decreased by at least 3.72, while DSC$\uparrow$ in SPIDER increased by at least 2.87\%. These factors suggest that integrating text prompts, even with basic category information, can significantly enhance ATM-Net's segmentation ability.

\textit{\textbf{(2) The finer the granularity of the text prompt, the better ATM-Net's performance, with Opt.3 yielding the best results.}}
With finer granularity, the selected text prompts provide more detailed anatomical descriptions and positional specifics, resulting in general improvements in ATM-Net's metrics. For instance, in MRSpineSeg, the Jaccard is 56.78 without text information, while it improves to 62.45, 66.84, and 68.46 for Opt.1, 2, and 3, respectively.
Specifically, Opt.3 prompts provide the most comprehensive information regarding image type, anatomical structures, and slice position, resulting in the best overall performance, as indicated in the last row of Table~\ref{tab:prompt-opt}.
% 在SPIDER上，边缘指标下降的原因。
It is noteworthy that, on the SPIDER dataset, Opt.2 achieved better boundary-aware metrics than Opt.3. We will discuss this issue in \Cref{sec:discuss}(3).

\section{Discussion}
\label{sec:discuss}

% This section discusses some key issues from the ATM-Net's design and experimental results, focusing on the impact of component integration, dataset-specific characteristics, and prompt optimization on ATM-Net’s segmentation performance.
In this section, we elaborate on the design principle of our prompt generation mechanism, discuss the generalizability of ATM-Net, and interpret key experimental findings.

% This section discusses the design principles of our prompt generation mechanism, the generalizability of ATM-Net, and interprets key experimental findings. 
% We first examine the explainable generation process and inherent flexibility of our dual-perspective prompts, then analyze component synergy and dataset-specific performance characteristics.

\subsection{Prompt Design and Explainability}
The text prompt generation mechanism is a critical component of ATM-Net, with two key issues requiring further discussion:

\noindent\textbf{\textit{Generation\,\&\,explainability.}} 
The generation process of our text prompt is explainable.
ATM-Net integrates dual-perspective text prompts:
(1) The holistic ones describe the overall information, including slice position and relative positions of subclasses.
(2) The class-wise, channel-level ones indicate presence/absence per subclass.
% ATM-Net uses prompts with two perspectives: a global perspective and a class-wise, channel-level perspective. The first describes the overall information of the image, including frame location and the relative positions of substructures. The latter clearly indicates whether each subclass is present in the image. 
Both prompts are automatically and adaptively obtained from image annotations ({\normalsize \textit{\Cref{fig:prompt-gen}}}), with extensive ablation studies validating their granularity impact ({\normalsize \textit{\Cref{tab:prompt-opt}}}).

% \noindent\textbf{\textit{Doctor interaction.}} 
% Doctors' prompts can take many forms, which can be very efficient:
% Akin to weak supervision solutions, inference prompts can be automatically generated via minimal clicks (\textit{e.g.}, sparse substructure presence cues).
\textcolor{black}{
\noindent\textbf{\textit{Inference mechanism \& doctor interaction.}} 
While ATPG utilizes mask annotations during training, ATM-Net does not use ground-truth mask information during inference. In practice, text prompts are generated via two practical mechanisms: (1) \textit{Fully automatic mode} using a fixed template that simply lists all possible structures. This alone achieves 75.02\% DSC on MRSpineSeg, which is a 10.44\% absolute improvement over the visual-only Swin UNETR baseline. 
(2) \textit{Interactive mode} requiring minimal clinician input: leveraging the regular sequential arrangement of lumbar structures, the clinician only needs to identify the first and last visible structures on a single reference slice per volume (e.g., T10 to L5). This information is readily available during routine examination and enables adaptive prompt generation with negligible additional effort. These mechanisms ensure both clinical applicability and fair comparison with pure vision models.
}

% \noindent\textbf{\textit{Prompt flexibility.}}
% In the main text, we validated that in HASF, the finer prompt granularity leads to better ATM-Net performance (\Cref{tab:prompt-opt}). 
% In addition, the channel-wise prompt in CCAE can also take other forms. Pixel-wise prompts like ``a pixel of VB T10'' can also improve the details of the segmentation. However, we chose class-wise, channel-level prompts that additionally improve inter-class discrimination with lower complexity.

\subsection{Further discussion on some experimental results}
\noindent\textbf{\textit{Component synergy in handling challenging cases.}}
In \Cref{tab:ablation}, ATM-Net gets comparative or slightly worse results in some boundary-aware metrics with only one of the components.
This phenomenon is because some of the comparing models fail to accurately segment three particularly challenging categories in the SPIDER dataset, leading to \textit{NaN (Not-a-Number)} values for ASD and HD95 metrics, which are then excluded from the overall evaluation process, leading to inflated results. 
When CCAE or HASF is used individually, these modules can identify but not always accurately segment the difficult regions, leading to slightly lower ASD and HD95 scores. However, when both CCAE and HASF are integrated, ATM-Net can handle these challenging cases more effectively, leading to better ASD and HD95 results.

\noindent\textbf{\textit{Dataset-specific performance differences.}}
ATM-Net's performance on SPIDER dataset is less optimal compared to the MRSpineSeg due to the following factors.
Firstly, the SPIDER dataset contains a significant amount of highlight noise.
% that the region-based mean cropping method mentioned earlier cannot fully eliminate. 
Secondly, accurately delineating spinal canal boundaries poses a substantial challenge unique to the SPIDER dataset but not present in the MRSpineSeg dataset. These complexities inherently increase the difficulty of achieving high segmentation accuracy on the SPIDER dataset.

\noindent\textbf{\textit{Prompt granularity optimization in noisy datasets.}} As shown in \Cref{tab:prompt-opt}, on the SPIDER dataset, Opt.2 achieved better boundary-aware metrics than Opt.3.
This is because the MRSpineSeg dataset consists of sagittal images without noise, whereas the SPIDER dataset contains highlight noise. In such cases, overly rich semantic information may hinder fine-grained segmentation, particularly in edge segmentation. Opt.2, with its moderate information granularity, reduces the impact of noise fitting, leading to relatively better performance.

\textcolor{black}{
\subsection{Generalizability to Other Anatomical Structures}
The proposed ATM-Net demonstrates strong potential for generalizability. Since the ATPG module is inherently modality-independent, it can easily adapt to other imaging modalities (\textit{e.g.}, CT, X-ray) and anatomical structures with consistent spatial topologies, such as the cervical spine or cardiac structures. However, applying this framework to highly variable organs (\textit{e.g.}, liver or tumors) remains challenging, as they lack the predictable spatial relationships required by our current sequential priors. To overcome this limitation, a promising future direction is to integrate real-world clinical reports into the prompt generation pipeline. This would enable the model to capture more complex, patient-specific anatomical variations, significantly extending its applicability to broader medical imaging tasks.
}

\section{Conclusion}
This study presents ATM-Net, an innovative framework that integrates anatomy-aware text guidance with multi-modal fusion for the fine-grained segmentation of the lumbar spine in MRI. Our method stands out due to its ability to generate informative text prompts in an annotation-free manner. It provides deep anatomical insights that are effectively integrated with image features, thus overcoming the various limitations of the existing solutions.

Our comprehensive experimental evaluations demonstrate that ATM-Net outperforms current SOTA methods and other VLMs across various metrics, especially regarding class discrimination and segmentation details. These results highlight the potential of our approach in providing clinicians with detailed, reliable segmentations that are pivotal for accurate diagnosis of spinal conditions.

Note that the ATPG module is annotation-free and modality-independent. Its design for adaptive text prompt generation and the integration with visual models can be easily transferred to other imaging modalities and various medical or non-medical tasks.
While our method shows impressive results, it has limitations. Currently, text prompts are generated only from image annotations, which may not cover broad anatomical knowledge.
Looking ahead, we see exciting research directions: (1) We plan to enhance ATPG by including more knowledge sources, like clinical reports, to improve text prompts. 
(2) We aim to fully leverage the transfer potential of ATM-Net by applying it to more imaging modalities and a wider range of application scenarios.

\section*{Acknowledgements}
This work is supported by the National Natural Science Foundation of China (No.62501160, 62173091, U24A20219 and 62272281),
The Fujian Provincial Natural Science Foundation of China (No.2023J05117, 2024J09021),
Special funds for Taishan Scholars Project under grant No. tsqn202306274.
The authors would also like to acknowledge the support from the Key Laboratory of Intelligent Metro of Universities in Fujian, Fuzhou University, Fuzhou, China, and the Fujian Key Laboratory of Network Computing and Intelligent Information Processing (Fuzhou University), Fuzhou, China.

\section*{Conflict of interests}
The authors declare that no potentail competing interests exist. There is no an undisclosed relationship thay may pose a competing interest. There is no an undisclosed funding source that may pose a competing interest.
	
\section*{Data availability}
Data will be made available on request.

\bibliography{sn-bibliography}% common bib file

\end{document}